\documentclass[10pt,twocolumn,letterpaper]{article}

\usepackage[pagenumbers]{cvpr} 
\usepackage{booktabs}
\usepackage[table]{xcolor}
\usepackage{multirow}
\usepackage{pifont}
\usepackage{comment}
\usepackage{annotate-equations}
\newcommand\awesomepose{{{PoseGAM}}\xspace}

\definecolor{cvprblue}{rgb}{0.21,0.49,0.74}
\usepackage[pagebackref,breaklinks,colorlinks,allcolors=cvprblue]{hyperref}

\def\ie{\textit{i.e.}}
\def\etc{\textit{etc}}
\def\eg{\textit{e.g.}}
\def\etal{\textit{et al.}}

\title{\awesomepose: Robust Unseen Object Pose Estimation via Geometry-Aware Multi-View Reasoning}

\author{Jianqi Chen~~~~~Biao Zhang~~~~~Xiangjun Tang~~~~~Peter Wonka\\
KAUST\\
{\tt\small \{jianqi.chen, biao.zhang, xiangjun.tang, peter.wonka\}@kaust.edu.sa}\\
\url{https://windvchen.github.io/PoseGAM}
}

\begin{document}
\twocolumn[{%
    \renewcommand\twocolumn[1][]{#1}%
    \maketitle
    \begin{center}
        \centering
        \captionsetup{type=figure}
        \includegraphics[width=\textwidth]{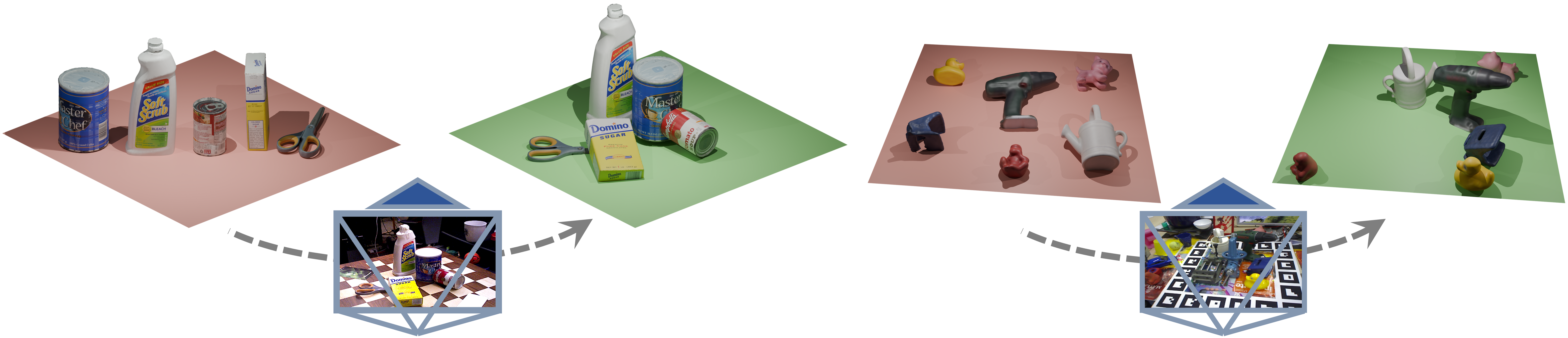}
        \vspace{-23pt}
        \captionof{figure}{\textbf{\awesomepose aligns CAD models with the query image.} Given the input CAD models, \awesomepose accurately estimates object poses that align with their spatial arrangements in the query image. Objects on the red plane indicate the initial CAD poses, while those on the green plane represent the poses after estimation.}
        \label{fig:teaser}
    \end{center}%
}]

\begin{abstract}
6D object pose estimation, which predicts the transformation of an object relative to the camera, remains challenging for unseen objects. Existing approaches typically rely on explicitly constructing feature correspondences between the query image and either the object model or template images. In this work, we propose \awesomepose, a geometry-aware multi-view framework that directly predicts object pose from a query image and multiple template images, eliminating the need for explicit matching. Built upon recent multi-view-based foundation model architectures, the method integrates object geometry information through two complementary mechanisms: explicit point-based geometry and learned features from geometry representation networks. In addition, we construct a large-scale synthetic dataset containing more than 190k objects under diverse environmental conditions to enhance robustness and generalization. Extensive evaluations across multiple benchmarks demonstrate our state-of-the-art performance, yielding an average AR improvement of 5.1$\%$ over prior methods and achieving up to 17.6$\%$ gains on individual datasets, indicating strong generalization to unseen objects.
\end{abstract}


\vspace{-20pt}
\section{Introduction}
\label{sec:intro}
\vspace{-5pt}
6D object pose estimation, \ie, predicting an object's rotation and translation relative to the camera coordinate system, has long been an important research topic with broad applications in robotic manipulation~\cite{okorn2021zephyr, sun2022ick}, augmented and virtual reality~\cite{su2019deep, li2024gbot}, autonomous driving~\cite{hoque2023deep, wu20196d}, content creation~\cite{chen2025v2m4, yenphraphai2025shapegen4d}, \etc. Early works primarily focused on instance-specific~\cite{xiang2017posecnn, peng2019pvnet} or category-specific~\cite{zhang2023generative, lin2024instance} object pose estimation. However, such approaches are limited in practical scenarios, as they often fail to generalize to objects unseen during training.

To improve generalization, recent research has shifted toward unseen object pose estimation. Existing methods typically follow either a \textit{match-then-localize}~\cite{zhao2023learning, nguyen2024gigapose, nguyen2024nope} or \textit{match-then-refine}~\cite{megapose, moon2024genflow, wen2024foundationpose, ornek2024foundpose, deng2025pos3r} paradigm. These methods explicitly establish feature correspondences between the query image and either the 3D model of the object or a set of template images with known poses. Once these correspondences are constructed, the object pose in the query image is recovered using standard geometric solvers such as least squares optimization or the PnP algorithm. Although these approaches have achieved promising results, their performance depends on the quality of the matching stage, which leads to pose estimation inaccuracies when matching is unreliable~\cite{liu2024deep}.

In this work, we explore whether unseen object pose estimation can be addressed through an end-to-end network, \textit{eliminating the need for explicit feature matching} and minimizing reliance on camera imaging priors. Inspired by recent multi-view foundation models~\cite{wang2025vggt, jiang2025rayzer, wang2025pi}, which have demonstrated the ability to directly infer 3D geometry without traditional structure-from-motion steps, we adopt and extend their architecture to the 6D pose estimation setting. Specifically, we design a multi-view model that jointly processes the query image and multiple template images with known poses, enabling the network to reason about the object’s pose directly.

Architecture inheritance enables us to exploit the powerful pretrained weights of multi-view foundation models and potentially extend their success to the object pose estimation task. However, existing multi-view foundation models rely solely on visual image inputs. Although they can estimate camera poses, they lack explicit information about the 3D object model, which is typically available in pose estimation tasks. This omission limits their effectiveness for object pose estimation. Moreover, these models typically assume appearance consistency across views, making them sensitive to appearance variations that frequently occur due to the domain gap between renderings of CAD models and real-world observations (see Appendix~\ref{app: fragile} for an analysis).

To address these limitations, we incorporate object geometry information into the multi-view architecture and construct a large-scale dataset for object-centric pose estimation. Specifically, for network design, we explore two complementary approaches: (1) injecting explicit point-based geometry and (2) integrating learned geometry features through a geometry representation network. We observe that directly feeding raw sequential geometry tokens into the multi-view structure hinders learning; therefore, we project geometry features back into view map representations, which better align with the model’s multi-view reasoning process. For dataset construction, to enhance robustness to object variations and visual inconsistencies, we build a large-scale and diverse synthetic dataset comprising over 190k objects with corresponding images under a wide range of challenging conditions, including variations in lighting, appearance, and other scene factors. This diversity enables our model to generalize effectively across different pose estimation scenarios.

Our main contributions are summarized as follows:

\begin{itemize}

\item We propose a multi-view feedforward network for object pose estimation. The network directly takes the query image and template images as input and predicts the object pose in an end-to-end manner, eliminating the explicit feature matching step used in prior works.

\item We introduce object geometry into the multi-view framework using explicit point maps and learned geometry representations. Instead of using raw geometry feature tokens, we project the features into view-map representations, enabling the network to reason more effectively about object poses and improving robustness across diverse scenarios.

\item We construct a large-scale and diverse synthetic object pose estimation dataset containing over 190k objects across multiple challenging scenarios, including varying environmental lighting conditions, appearance variations, and other real-world complexities.

\end{itemize}

\section{Related Works}
\label{sec: Related Works}

\subsection{Object Pose Estimation}

Object pose estimation aims to determine an object's transformation relative to the camera, typically given an observed image and the object’s geometric model. Traditional methods can be broadly categorized into instance-level and category-level approaches. Instance-level methods~\cite{li2019cdpn, rad2017bb8, li2022dcl, sundermeyer2018implicit, wang2019densefusion, peng2019pvnet, xiang2017posecnn, hu2020single} are designed or trained for a specific object instance. By employing techniques such as correspondence prediction~\cite{li2019cdpn, rad2017bb8}, template matching~\cite{li2022dcl, sundermeyer2018implicit}, keypoint voting~\cite{wang2019densefusion, peng2019pvnet}, or direct pose regression~\cite{xiang2017posecnn, hu2020single}, these methods can achieve highly accurate pose estimation. However, their applicability is limited, as each new object instance requires retraining or fine-tuning. This limitation has motivated the development of category-level methods~\cite{tian2020shape, wang2021category, irshad2022centersnap, lin2022sar, wang2019normalized, fan2022object, chen2021fs, wang20196-pack}, which seek to generalize across unseen objects within the same category. Many such methods~\cite{tian2020shape, wang2021category, fan2022object} first extract a category-specific shape prior and then align the query object to this canonical shape before estimating its pose. Other works~\cite{chen2021fs, wang20196-pack} attempt to directly regress the pose without explicitly modeling the shape prior. Although these approaches improve generalization within a category, they still struggle to handle the diversity of real-world object appearances. Recently, research has shifted toward unseen object pose estimation~\cite{chen2023zeropose, huang2024matchu, lin2024sam, nguyen2024gigapose, megapose, geng2025one}, where the goal is to estimate the poses of category-agnostic novel objects. These methods typically train networks to extract representative features from both geometry and images, and then derive the pose through cross-modal correspondences. Some works~\cite{ornek2024foundpose, ausserlechner2024zs6d, caraffa2024freeze} further exploit powerful pretrained feature extractors to obtain these representations directly.

\begin{figure*}[!t]
  \centering
  \includegraphics[width=\linewidth]{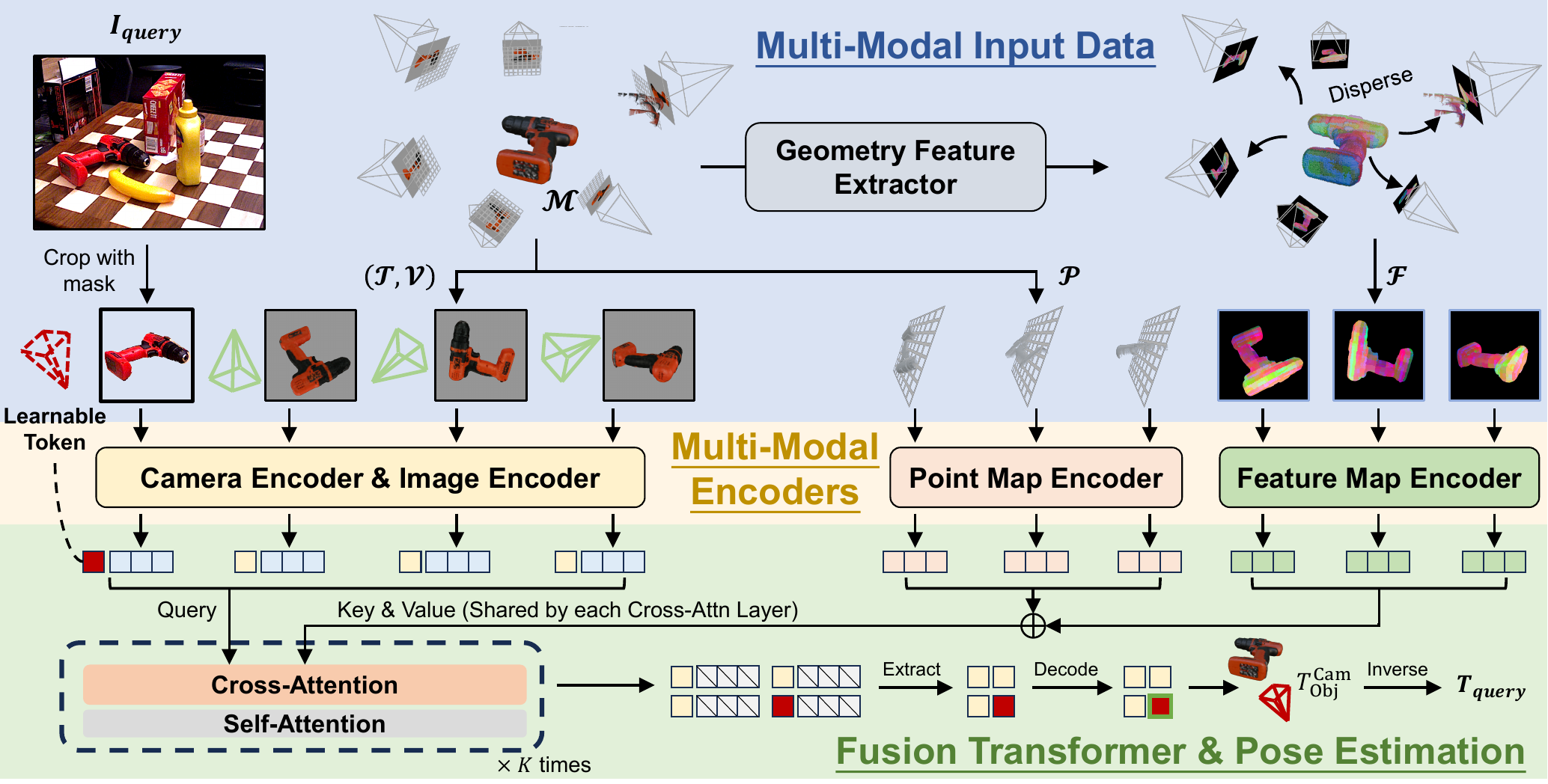}
  \vspace{-20pt}
  \caption{\textbf{Overview of \awesomepose.} Given a query image $I_{\text{query}}$ and an object mesh $\mathcal{M}$, the goal is to estimate the object-to-camera transformation ${T}_{\text{query}}$. A set of camera poses is sampled around $\mathcal{M}$ to render images $\mathcal{V}$ and corresponding point maps $\mathcal{P}$. Both $I_{\text{query}}$ (with its foreground segmented) and $\mathcal{V}$ are encoded into image tokens, each paired with a camera token. For the rendered views $\mathcal{V}$, the camera tokens are computed from known intrinsics and extrinsics, whereas for $I_{\text{query}}$ the camera token is a learnable embedding. A geometry feature extractor produces a global object representation, which is distributed to camera views to form view-specific features $\mathcal{F}$. These, together with point maps $\mathcal{P}$, are encoded as key–value tokens for cross-attention. The output camera tokens are decoded to predict the camera-to-object transformation ${T}^{\text{Cam}}_{\text{Obj}}$, from which the final object-to-camera pose $T_{\text{query}}$ is obtained by matrix inversion.}
  \label{fig:framework}
  \vspace{-10pt}
\end{figure*}

\subsection{Multi-View Foundation Models}

Traditional geometric reasoning methods~\cite{schonberger2016structure, cui2017hsfm, crandall2012sfm}, which reconstruct sparse 3D maps while jointly estimating camera parameters from multiple images, typically rely on Structure-from-Motion (SfM). In these pipelines, pixel correspondences are first obtained through keypoint matching across images to establish geometric relationships, followed by bundle adjustment to jointly optimize 3D coordinates and camera parameters. Dense geometry can then be reconstructed using Multi-View Stereo (MVS)~\cite{goesele2006multi}. To avoid such complex intermediate steps, recent approaches aim to predict 3D geometry directly from RGB images. Since single-image reconstruction is inherently ill-posed, these methods employ neural networks trained on large datasets to learn strong 3D priors, helping resolve ambiguities. For example, DUSt3R~\cite{wang2024dust3r} and its metric follow-up MASt3R~\cite{leroy2024grounding} predict relative point maps from image pairs. Additional scene representations, such as camera poses and depth, can then be recovered by iteratively processing multiple image pairs and applying post-optimization. Subsequent works like VGGT~\cite{wang2025vggt} and $\pi^3$~\cite{wang2025pi} extend DUSt3R to multi-view settings. VGGT constructs a multi-image network that employs an alternating-attention transformer to predict multi-view point maps, depth, camera poses, and tracking features. $\pi^3$ further refines VGGT by removing the need for the first input frame as a reference coordinate. Similarly, RayZer~\cite{jiang2025rayzer} processes unposed and uncalibrated multi-view images to predict per-view camera parameters along with a consistent scene representation.

\subsection{Ours versus Others}

Unlike existing unseen object pose estimation methods, which rely on explicitly constructing feature correspondences, our approach employs a multi-view network inspired by the success of such architectures in recent geometric reasoning fields. The network directly predicts the object pose by jointly processing the query image and multiple object template images as input. In contrast to typical multi-view foundation models, we incorporate the object’s geometric information into the network, which enhances both precision and accuracy, making the architecture well-suited for the object pose estimation task.

\section{Method}
\label{sec: Method}
Given an object $\mathcal{M}$ and a query image $I_{\text{query}}$ containing this object, our goal is to estimate the object pose $T_{\text{query}}$ with respect to the camera coordinate system:
\begin{equation}
T_{\text{query}} = \mathrm{Network}(I_{\text{query}}; \mathcal{M}).
\end{equation}

Recent multi-view foundation models, such as VGGT~\cite{wang2025vggt}, $\pi^3$~\cite{wang2025pi}, and RayZer~\cite{jiang2025rayzer}, have achieved remarkable results in geometric reasoning tasks. Motivated by their success, we incorporate multi-view information into our framework. This design offers two key advantages: 1) we can leverage recent successful architectures of these recent multi-view networks; 2) we can initialize our model using large-scale pretrained weights, facilitating faster convergence and improved generalization. 

Specifically, we render a set of multi-view RGB images $\mathcal{V}=\{I_i\}_{i=1, \cdots, N}$ of the object $\mathcal{M}$ using manually defined camera transformations $\mathcal{T}=\{T_i\}_{i=1, \cdots, N}$ around object $\mathcal{M}$. The queried pose is then estimated as:
\begin{equation}\label{eq:image-model}
    T_{\text{query}} = \mathrm{Network}(I_{\text{query}};\mathcal{T}, \mathcal{V})
\end{equation}
The main multi-view RGB image network is detailed in Sec.~\ref{sec:rgb-view}.

Furthermore, the inputs can be augmented with additional geometric information derived from the known object model $\mathcal{M}$, leading to the final formulation of our network:
\vspace{20pt}
\begin{equation}\label{eq:multi-model}
\hspace{-5em}
    T_{\text{query}} = \mathrm{Network}(I_{\text{query}};
        \eqnmark{nodeT}{\mathcal{T}},
        \eqnmark{nodeV}{\mathcal{V}},
        \eqnmark{nodeP}{\mathcal{P}},
        \eqnmark{nodeF}{\mathcal{F}}
        )
        \annotate[yshift=-0.5em]{below,left}{nodeT}{transformations (cameras)}
        \annotate[yshift=0.5em]{above,left}{nodeV}{multi-view RGB images}
        \annotate[yshift=-0.5em]{below, right}{nodeP}{point maps}
        \annotate[yshift=0.5em]{above, right}{nodeF}{point cloud features}
        \vspace{20pt}
\end{equation}
where $\mathcal{P}$ denotes the point maps (Sec.~\ref{sec:point-map}) and $\mathcal{F}$ represents the corresponding per-point features (Sec.~\ref{sec:point-net}).
The overall pipeline of our method is illustrated in Fig.~\ref{fig:framework}.

\subsection{Multi-View Network}\label{sec:rgb-view}
The main network takes camera poses $\mathcal{T}$ and the corresponding multiview RGB images $\mathcal{V}$ as input. To process this multiview information, we first extract feature tokens $X_i = (\mathbf{x}_i^{(1)}, \mathbf{x}_i^{(2)}, \cdots, \mathbf{x}_i^{(L)})$ for each image $I_i$ using a pretrained network (DINOv2~\cite{oquab2023dinov2}). Concurrently, we encode camera poses $T_i$ into a dedicated camera token $\mathbf{c}_i$ using a lightweight camera encoder:
\begin{equation}
    (\mathcal{V}, \mathcal{T})
    \rightarrow 
        \{
            (\mathbf{x}_i^{(1)}, \mathbf{x}_i^{(2)}, \cdots, \mathbf{x}_i^{(L)}, \mathbf{c}_i)
        \}_{i=1, \cdots, N}
\end{equation}
Similarly, for the query image $I_{\text{query}}$, we extract its visual tokens $X_{\text{query}}$ and append a learnable query camera token $\mathbf{c}_{\text{query}}$. The total number of tokens processed by the network is therefore:
\begin{equation}
    \overbrace{
        \underbrace{
            (L+1)
        }_{\text{intra-frame}}
        \times(N+1)
    }^{\text{inter-frame}}
\end{equation}

Our network architecture is inspired by the design of VGGT~\cite{wang2025vggt}, a feed-forward multi-view foundation model. We alternately apply inter- and intra-frame self-attention layers to these tokens (named \textbf{multiview tokens}). Furthermore, we inject \textbf{geometry tokens} into the main network to incorporate explicit 3D information. Specifically, before each self-attention layer, we perform a cross-attention operation, denoted as $\mathrm{CA}$, between the multi-view tokens and the geometry tokens:
\begin{equation}\label{eq:cross-attn}
    \mathrm{CA}(Q\leftarrow\textbf{multiview tokens}, KV\leftarrow\textbf{geometry tokens})
\end{equation}
The processing and construction of these geometry tokens are detailed in Sec.~\ref{sec:point-map} and Sec.~\ref{sec:point-net}.

After passing through multiple attention layers, the features corresponding to the camera tokens are decoded by a lightweight head to predict the camera pose. The resulting output is supervised solely by the ground-truth camera pose.

\subsection{Geometry Processing via Point Maps}\label{sec:point-map}
We process the object geometry $\mathcal{M}$ through the following steps. First, we render the object into multi-view depth maps using the camera poses $\mathcal{T}$, and subsequently reconstruct the point maps in world coordinates using the corresponding camera intrinsics:
\begin{equation}\label{eq:point-map}
    \text{Object } \mathcal{M}\overset{\mathcal{T}}{\rightarrow} \text{Depth Maps} \overset{}{\rightarrow} \text{Point Maps } \mathcal{P}
\end{equation}
where $\mathcal{P}=\{P_i\}_{i=1,\cdots, N}$. 
Each point map $P_i$ is processed by a lightweight convolutional neural network to produce a set of point map tokens:
\begin{equation}
P_i\overset{\mathrm{Conv}}{\longrightarrow}\left\{\mathbf{p}_i^{(1)}, \mathbf{p}_i^{(2)}, \cdots, \mathbf{p}_i^{(L)}\right\}
\end{equation}
A naive approach would be to directly add the point map tokens to the multi-view RGB image tokens. However, such direct fusion introduces a substantial modality gap from the pretrained model’s original input distribution (which is primarily based on natural images), thereby hindering effective knowledge transfer (see Sec.~\ref{sec: Experiments}). To mitigate this, we employ a cross-attention mechanism for geometry information injection, as defined in Eq.~\eqref{eq:cross-attn}.

\subsection{Geometry Processing via Point Cloud Networks}\label{sec:point-net}
To further enhance the network's understanding of the object model, we employ off-the-shelf geometry representation networks to extract a global representation of the 3D object and inject it into our framework. Specifically, we adopt existing point cloud architectures (\eg, PointTransformer v3~\cite{wu2024ptv3}). The input to the network is a point cloud (with coordinates recovered from Eq.~\eqref{eq:point-map}) augmented with per-point color and normal information. The network outputs a set of per-point feature embeddings. In practice, we observe that directly injecting these features in their raw format is ineffective (see Sec.~\ref{subsec: exp geo}); the model struggles to utilize the information efficiently. Instead, when the per-point features are spatially reorganized into a view-map format, the network can more effectively leverage the encoded geometric structure. Motivated by this observation, we replace the coordinate channels in the point maps with the extracted feature vectors, thereby forming feature maps:
\begin{equation}
    \text{Point Clouds}\overset{\text{PCNet}}{\longrightarrow}\text{Per-Point Features}\overset{\text{Disperse}}{\longrightarrow}\text{Feature Maps }\mathcal{F}
\end{equation}
where $\mathcal{F}=\{F_i\}_{i=1,\cdots, N}$. 
Analogous to point maps, we apply a lightweight convolution network to these feature maps to obtain feature tokens:
\begin{equation}
F_i\overset{\mathrm{Conv}}{\longrightarrow}\left\{\mathbf{f}_i^{(1)}, \mathbf{f}_i^{(2)}, \cdots, \mathbf{f}_i^{(L)}\right\}
\end{equation}
These feature tokens are then added to the corresponding point map tokens, jointly forming the $KV$ inputs in the cross-attention layers.

\section{Data Construction}
\label{sec: Data Construction}

To ensure the robustness of our method for object pose estimation, we construct a large-scale synthetic dataset comprising a diverse collection of 3D objects. Each object ($\mathcal{M}$) is paired with texture-rendered images ($\mathcal{V}$) and geometry-related maps ($\mathcal{P}$) captured under a wide range of camera poses ($\mathcal{T}$). The overall data construction pipeline consists of two stages: geometry data preparation and image/map generation. Additional details of the dataset, along with ablation studies validating its effectiveness, are provided in Appendix~\ref{app: dataset}~\&~\ref{app: more quantitative}.

\begin{figure}[!t]
  \centering
  \includegraphics[width=\linewidth]{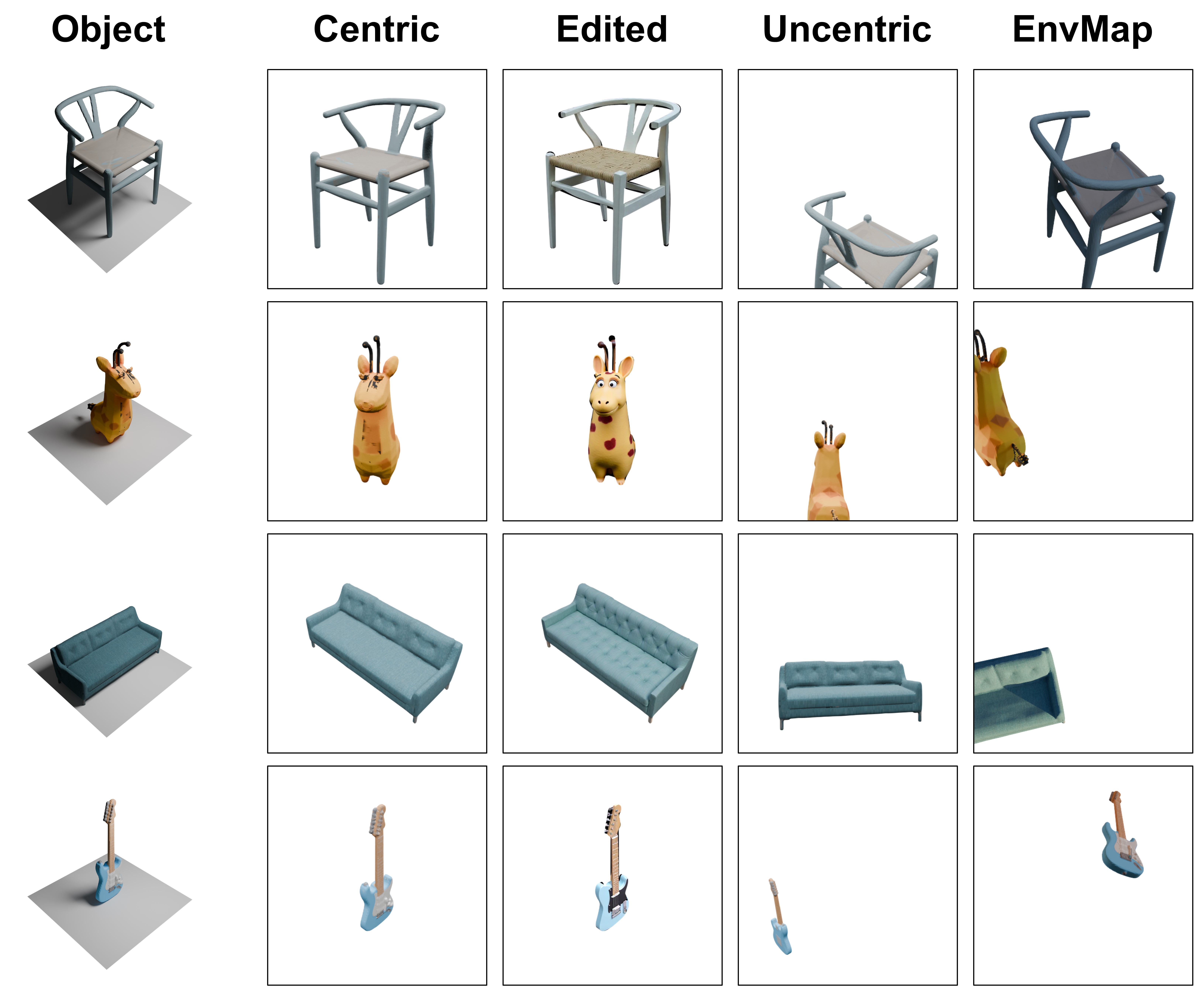}
  \vspace{-20pt}
  \caption{\textbf{Examples from the constructed object pose estimation dataset.} The leftmost column shows the object mesh after texture rebaking. The four columns on the right illustrate the four types of rendered image data.}
  \label{fig:dataset}
\vspace{-10pt}
\end{figure}

\subsection{Geometry Data Construction}

We collect synthetic 3D object assets from multiple publicly available datasets, including Toys4K~\cite{stojanov2021using}, 3D-FUTURE~\cite{fu20213d}, ABO~\cite{collins2022abo}, HSSD~\cite{khanna2024habitat}, and Objaverse~\cite{deitke2023objaverse}. To ensure geometric integrity and overall asset quality, we follow the filtering strategy of Xiang~\etal~\cite{xiang2025structured}, which removes objects with low-quality geometry or unrealistic mesh structures. After filtering, we retain over 190,000 high-quality object assets.

Following the asset selection process, we further standardize the objects to ensure consistency during rendering. Many assets contain complex shader graphs or procedural materials, which can lead to undesirable variations across different rendering platforms. To address this, we perform texture re-baking. Specifically, we apply Smart UV Unwrap in Blender~\cite{blender} to obtain consistent UV coordinates, and then bake a consolidated texture that integrates the Diffuse, Glossy, and Transmission components. This procedure eliminates shader-dependent inconsistencies and produces a uniform material representation for all assets. Finally, each object is exported in GLB format with a single base color texture map.

\subsection{Image Data Construction}
\label{subsec: image data}

For each object asset, we define 50 camera poses distributed uniformly around the object using a spherical Hammersley sequence. At each pose, we render both the texture image and a set of geometry-related maps, including depth maps, normal maps, and mask maps. These rendered modalities serve as the geometric inputs required by our model (see Sec.~\ref{sec: Method}). To prepare the query images, we render each asset under four distinct scenarios designed to introduce varying levels of difficulty. This diversity helps strengthen the model’s robustness under realistic conditions. Examples of the resulting dataset are shown in Fig.~\ref{fig:dataset}; for clarity, only RGB renderings are partially displayed.

\vspace{-10pt}
\paragraph{\textbf{Centric Object Images.}} In the first scenario, camera poses are uniformly sampled around the object, and all viewing directions point toward its centroid. The lighting environment is fixed and consists of three sources: a top area light, a bottom area light, and a front-top point light. Rendering is performed using Blender EEVEE~\cite{blender}, which offers a favorable balance between visual fidelity and computational efficiency. This configuration represents the most basic case, where the object remains centered in the frame and is observed under consistent illumination.

\vspace{-10pt}
\paragraph{\textbf{Uncentric Object Images.}} To simulate more natural image compositions in which the object does not appear at the center of the frame, we randomize both the camera position and the corresponding look-at point. Starting from the centric configuration, we first sample camera poses on the viewing sphere and then apply random perturbations to both the viewpoint and the orientation. For each object, multiple pose candidates are tested until a valid pose is identified, which we define as one where more than 30$\%$ of the object’s vertices are projected within the image plane. This procedure yields images with diverse object placements and varying degrees of visibility, thereby improving robustness to off-center viewpoints and partial occlusions.

\begin{figure*}[!t]
  \centering
  \includegraphics[width=\linewidth]{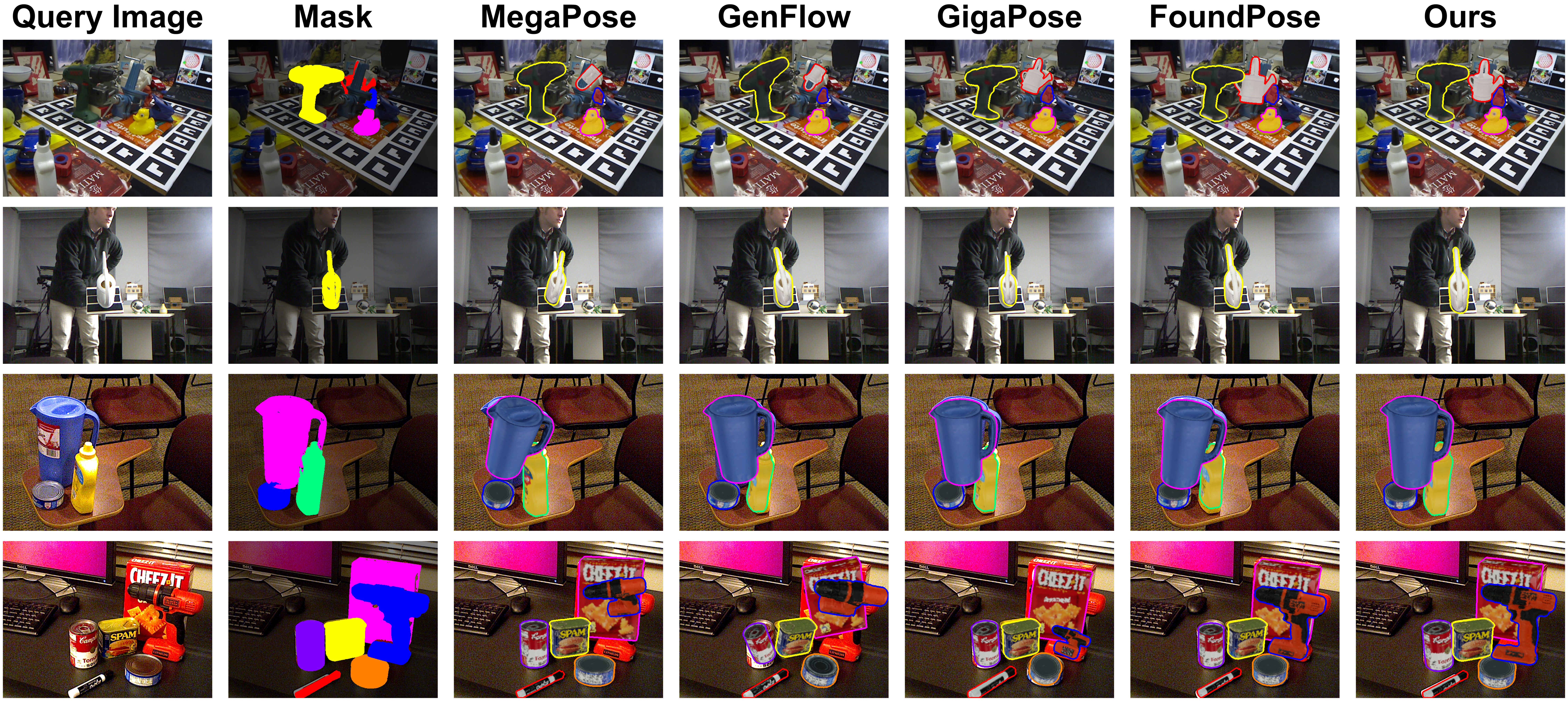}
  \vspace{-20pt}
  \caption{\textbf{Visual comparison with other methods.} From left to right: the query image for pose estimation, object masks with each object shown in a distinct color, and the projection results of different methods after applying the estimated poses to the 3D object models. The projected 3D models are outlined with borders matching the colors of their corresponding masks.}
  \label{fig:comparison}
  \vspace{-15pt}
\end{figure*}

\vspace{-10pt}
\paragraph{\textbf{Uncentric Object Images with Varying Lighting.}} To further enhance appearance diversity and approximate realistic illumination conditions, we incorporate more than 800 HDR environment maps from Poly Haven~\cite{polyheaven}. For each rendering, one HDR map is randomly selected to define the surrounding lighting environment. Rendering is performed using Blender Cycles~\cite{blender}, which provides physically accurate light transport and reflections. This scenario introduces substantial variation in shading, color temperature, and specular behavior, enabling the model to generalize more effectively to real-world lighting.

\vspace{-10pt}
\paragraph{\textbf{Centric and Uncentric Objects with Edited Appearances.}} To address challenging cases in which the query image presents appearance inconsistencies relative to the reference object while still preserving structural alignment, we additionally generate a set of appearance-edited images. Such inconsistencies can degrade the performance of multi-view foundation models (see Appendix~\ref{app: fragile} for analysis), which often assume appearance-consistent viewpoints. To introduce controlled appearance variations, we employ DDIM inversion~\cite{song2020denoising}, a technique widely used in image editing~\cite{mokady2023null, chen2025zero}. Specifically, we inject Gaussian noise into the centric and uncentric renderings at multiple noise levels and then denoise them using FLUX~\cite{flux1_canny_dev}, a text-conditioned diffusion model. During the denoising stage, the model is conditioned on the object descriptions provided by Xiang~\etal~\cite{xiang2025structured}.

This process preserves the underlying geometry of the object while producing diversified textures and appearances across the samples. To ensure structural consistency and avoid invalid generations, such as synthesizing a front-facing appearance from a noisy backside image, we restrict appearance editing to camera poses with pitch angles between 15$^{\circ}$ and 60$^{\circ}$ and yaw angles within $\pm$60$^{\circ}$. We further discard any generated images in which the object extends beyond the image boundaries.

\section{Experiments}
\label{sec: Experiments}

Please refer to Appendix~\ref{app: implement} for implementation details, including the network architecture, training, inference, and evaluation settings. Additional quantitative results, including ablation studies of our dataset, further comparisons with existing methods, and visual comparisons across diverse scenarios, are provided in Appendix~\ref{app: more quantitative} and~\ref{app: visual}.

\subsection{Experimental Setup}

\paragraph{\textbf{Training and Evaluation Datasets.}} We train our model on the synthetic dataset constructed as described in Sec.~\ref{sec: Data Construction}. To evaluate its performance on unseen object pose estimation, we benchmark our approach on five widely used datasets: LM-O~\cite{brachmann2014learning}, T-LESS~\cite{hodan2017tless}, YCB-V~\cite{xiang2018posecnn}, TUD-L~\cite{hodan2018bop}, and IC-BIN~\cite{doumanoglou2016recovering}. For these benchmark experiments, the model is trained on the full constructed dataset to ensure comprehensive learning. In contrast, for ablation studies, training and evaluation are performed on a subsampled version of the dataset, which enables efficient experimentation.

\begin{table*}[t]
    \centering
    \resizebox{0.8\linewidth}{!}{%
    \begin{tabular}{lcccccc}
        \toprule
        \textbf{Method} & \textbf{LM-O~\cite{brachmann2014learning}} & \textbf{T-LESS~\cite{hodan2017tless}} & \textbf{TUD-L~\cite{hodan2018bop}} & \textbf{IC-BIN~\cite{doumanoglou2016recovering}} & \textbf{YCB-V~\cite{xiang2018posecnn}} & \textbf{Average}\\
        \midrule
        OSOP~\cite{osop}       & 31.2 & --   & --   & --   & 33.2 & 32.2 \\
        ZS6D~\cite{ausserlechner2024zs6d}       & 29.8 & 21.0 & --   & --   & 32.4 & 27.7 \\
        MegaPose~\cite{megapose}   & 22.9 & 17.7 & 25.8 & 15.2 & 28.1 & 21.9 \\
        GenFlow~\cite{moon2024genflow}    & 25.0 & 21.5 & 30.0 & 16.8 & 27.7 & 24.2 \\
        GigaPose~\cite{nguyen2024gigapose}   & 29.9 & 27.3 & 30.2 & 23.1 & 29.0 & 27.9 \\
        FoundPose~\cite{ornek2024foundpose}  & 39.6 & 33.8 & 46.7 & \underline{23.9} & 45.2 & 37.8 \\
        RayPose~\cite{huang2025raypose}       & \underline{42.1}  & \cellcolor[HTML]{EFEFEF}\textbf{36.9} & \underline{48.3} & 21.8 & \underline{46.2} & \underline{39.1} \\
        VGGT~\cite{wang2025vggt}       & 10.6  & 12.8 & 30.0 & 14.5 & 13.5 & 16.3 \\
        Ours       &    \cellcolor[HTML]{EFEFEF}\textbf{43.0} & \underline{34.1} &  \cellcolor[HTML]{EFEFEF}\textbf{56.8}& \cellcolor[HTML]{EFEFEF}\textbf{24.3} &  \cellcolor[HTML]{EFEFEF}\textbf{47.4}& \cellcolor[HTML]{EFEFEF}\textbf{41.1}\\
        \bottomrule
    \end{tabular}
    }  
    \vspace{-5pt}
    \caption{\textbf{Performance Comparisons on BOP Datasets.} We report Average Recall (AR) scores on five BOP core datasets. All methods are evaluated without using refinement networks or multi-hypothesis strategies~\cite{megapose, moon2024genflow}. The best results are shown in \textbf{bold}, and the second-best results are \underline{underlined}. For additional comparisons, including performance with refinement networks, the number of reference images, and runtime cost, please refer to Appendix~\ref{app: more quantitative}.}
      \vspace{-15pt}
    \label{tab:compare}
\end{table*}

\vspace{-15pt}
\paragraph{\textbf{Evaluation Metrics.}} To compare our method with existing object pose estimation approaches, we adopt the standard Average Recall (AR) metric from the Benchmark for Pose Estimation (BOP)~\cite{hodavn2020bop}. AR is computed using three pose-error functions: Visible Surface Discrepancy (VSD), Maximum Symmetry-Aware Surface Distance (MSSD), and Maximum Symmetry-Aware Projection Distance (MSPD). A pose is considered correct if its error falls below a predefined threshold. The mean recall is calculated for each error function across multiple thresholds, and the overall AR is defined as: $AR = (AR_{\text{VSD}} + AR_{\text{MSSD}} + AR_{\text{MSPD}})/{3}$. For the ablation studies, we adopt a variant of the standard AUC@N metric~\cite{wang2025vggt}, which combines Absolute Rotation Accuracy (ARA) and Absolute Translation Accuracy (ATA). ARA and ATA measure the angular errors in rotation and translation, respectively, for each query image. These errors are thresholded to compute per-threshold accuracy scores, and the AUC is then calculated as the area under the curve of the minimum values between ARA and ATA across all thresholds.

\subsection{Comparisons}
\label{subsec: compare}
We evaluate our method on five core BOP benchmark datasets, which are unseen during training. We compare it with recent state-of-the-art matching-based methods~\cite{osop, ausserlechner2024zs6d, megapose, moon2024genflow, nguyen2024gigapose, ornek2024foundpose}, as well as the recent feedforward method RayPose~\cite{huang2025raypose}, which leverages multi-view diffusion. All methods use the same base information, namely the CAD model and object detection results (\ie, segmentation masks), and operate on RGB-only query images. As shown in Table~\ref{tab:compare}, our method achieves the highest average AR across all datasets, outperforming the previous best approach by 5.1$\%$ on average. Notably, on the TUD-L dataset, our method achieves a 17.6$\%$ improvement. We attribute this substantial gain to the fact that TUD-L consists of single-object images with minimal occlusion, which closely aligns with the conditions of our synthetic training data. Fig.~\ref{fig:teaser} and Fig.~\ref{fig:comparison} illustrate the pose estimation visualizations of our methods and others, further validating the robustness of our approach. These results collectively highlight the strong generalization capability of our method across diverse datasets.

We also include VGGT~\cite{wang2025vggt} for comparison. Since VGGT predicts only relative transformations, we adapt it for object pose estimation by providing images rendered from known poses and applying the predicted relative transformation to the corresponding rendering poses. The results indicate that initial multi-view foundation models struggle to generalize to the object pose estimation task. This limitation is primarily due to their lack of explicit object geometry information and sensitivity to appearance inconsistencies between CAD renderings and real-world images (see Appendix~~\ref{app: fragile} for further analysis). In contrast, our method, by incorporating geometry features and fine-tuning on a carefully constructed synthetic dataset, achieves robust performance in object pose estimation.

\vspace{-2pt}
\subsection{Ablation Study}
\label{subsec: ablate}
\vspace{-3pt}
\paragraph{\textbf{Input modality.}} Our final network architecture takes multiple input modalities. In addition to the basic RGB images $\mathcal{V}$, it also utilizes camera poses $\mathcal{T}$, point maps $\mathcal{P}$, and geometric feature maps $\mathcal{F}$. Table~\ref{tab:ablate modality} presents an ablation study on the impact of these modalities. Specifically, when only the camera pose is provided, the network performance degrades significantly. Incorporating point maps notably improves pose estimation accuracy, as reflected by higher AUC values under lower thresholds (\eg, @3 and @5). Further introducing geometric feature maps leads to an additional performance gain, demonstrating the effectiveness of injecting geometric information into the network. We also experimented with including more geometry-related data, such as depth maps (denoted as $\mathcal{D}$). However, this yielded limited improvement, likely due to information redundancy, as the network can already infer depth cues from the camera and point map inputs.

\begin{table}[t]
\centering
\resizebox{\linewidth}{!}{%
\begin{tabular}{@{}ccccc@{}}
\toprule
Input Modality & AUC@3$\uparrow$ & AUC@5$\uparrow$ & AUC@10$\uparrow$ & AUC@30$\uparrow$ \\ \midrule
$\mathcal{V}$+$\mathcal{T}$                             &  15.07 &  29.72 &  53.62 & 77.77 \\
$\mathcal{V}$+$\mathcal{T}$+$\mathcal{P}$               &  25.94 &  41.63 &  61.24 & 80.26 \\
$\mathcal{V}$+$\mathcal{P}$+$\mathcal{F}$               &  25.56 &  41.52 &  61.57 & 82.04 \\
$\mathcal{V}$+$\mathcal{T}$+$\{\mathcal{D,P}\}$+$\mathcal{F}$ &  \cellcolor[HTML]{EFEFEF}\textbf{28.75} &  \cellcolor[HTML]{EFEFEF}\textbf{45.60} &  65.16 & 84.20                     \\
$\mathcal{V}$+$\mathcal{T}$+$\mathcal{P}$+$\mathcal{F}$ &  28.18 &  45.51 &  \cellcolor[HTML]{EFEFEF}\textbf{65.31} & \cellcolor[HTML]{EFEFEF}\textbf{84.30}                     \\
\bottomrule
\end{tabular}}
\vspace{-5pt}
\caption{\textbf{Ablation of input modalities.} Comparison of network performance using different combinations of input modalities, including RGB images ($\mathcal{V}$), camera poses ($\mathcal{T}$), point maps ($\mathcal{P}$), depth maps ($\mathcal{D}$), and geometric feature maps ($\mathcal{F}$).}
\label{tab:ablate modality}
  \vspace{-5pt}

\end{table}

\begin{table}[t]
\centering
\resizebox{\linewidth}{!}{%
\begin{tabular}{@{}ccccc@{}}
\toprule
View Strategy & AUC@3$\uparrow$ & AUC@5$\uparrow$ & AUC@10$\uparrow$ & \multicolumn{1}{c}{AUC@30$\uparrow$} \\ \midrule
Random Sample &  14.23&  31.63&  53.29&  76.57                    \\
FPS Sample &  \cellcolor[HTML]{EFEFEF}\textbf{28.18} &  \cellcolor[HTML]{EFEFEF}\textbf{45.51} &  \cellcolor[HTML]{EFEFEF}\textbf{65.31} & \cellcolor[HTML]{EFEFEF}\textbf{84.30}                      \\ \midrule
$|\mathcal{T}|=5$                & 11.46&  25.54&  48.15&  76.36 \\
$|\mathcal{T}|=10$                 & {28.18} &  {45.51} &  {65.31} & {84.30}                    \\
$|\mathcal{T}|=20$   &  \cellcolor[HTML]{EFEFEF}\textbf{33.59} &  \cellcolor[HTML]{EFEFEF}\textbf{49.47}&  \cellcolor[HTML]{EFEFEF}\textbf{69.73}&  \cellcolor[HTML]{EFEFEF}\textbf{85.61}                      \\ 
\bottomrule
\end{tabular}}
\vspace{-5pt}
\caption{\textbf{Effect of view sampling strategy and number of views.} Performance comparison across different strategies for camera poses sampling and varying numbers of rendered images.}
  \vspace{-15pt}
\label{tab:ablate view}
\end{table}

\vspace{-10pt}
\paragraph{\textbf{Effect of view sampling strategy and number of views.}} Table~\ref{tab:ablate view} compares the performance of different strategies for sampling the known camera poses $\mathcal{T}$ used to render template images (as described in Sec.~\ref{sec: Method}). In the \textit{random sampling} approach, camera poses are selected randomly on a sphere, whereas \textit{farthest point sampling} (FPS) selects poses to maximize the distance between each sampled viewpoint. We observe that random sampling results in substantially lower performance. This is primarily because, unlike FPS, random sampling often fails to provide sufficient coverage of all object aspects, leading to incomplete object understanding. This conclusion is further supported by our analysis of the number of rendered views: as the number of sampled views increases, offering more comprehensive coverage of the object, pose estimation accuracy improves.

\vspace{-10pt}
\paragraph{\textbf{Effect of finetuning strategy.}} Table~\ref{tab:ablate finetune} evaluates the impact of different training strategies. Specifically, we compare training with and without initializing from the weights of existing multi-view foundation models. Under the same training budget, initializing from VGGT weights significantly accelerates network convergence. Additionally, for the geometric representation network, finetuning PTv3 yields better performance compared to freezing its pretrained weights. We attribute this to the ability of finetuning that better aligns geometric features with the rest of the network, enabling more effective fusion with other modalities.

\subsection{Analysis of geometry feature injection}
\label{subsec: exp geo}

We conducted several experiments to investigate the optimal way to extract global geometry representations and inject them into the network. A straightforward approach is to directly feed the extracted geometry features into the network. We evaluated this strategy using the VecSet~\cite{zhang20233dshape2vecset} and PTv3~\cite{wu2024ptv3} representations. For VecSet, we employed pretrained weights from Step1X-3D~\cite{li2025step1x}, while for PTv3, we used weights from Sonata~\cite{wu2025sonata}. 

The output geometry representation from VecSet is a sequence of tokens, whose length corresponds to a downsampled version of the original object model's vertices. In contrast, PTv3 outputs point features with a number equal to the input vertices. We initially injected these features in their raw format ($\mathbb{R}^{L\times C}$) into the network via a separate cross-attention layer. However, as shown in Table~\ref{tab:ablate geometry injection}, this direct injection underperforms compared to first projecting the features into a view-based format ($\mathbb{R}^{N\times H\times W\times C}$) before fusion. This indicates that the network benefits from handling geometry features in a view-map representation, which better aligns with the visual input.

We further examined the role of color information in geometry features. Incorporating texture color into VecSet improves precision by about 3 points, likely because geometric features enriched with color are better aligned with RGB image features during attention. Moreover, directly adding geometry features to RGB features disrupts the pretrained model’s input distribution, leading to training instability and near-collapse of the network. These findings highlight the importance of our cross-attention injection strategy for effectively integrating geometry representations.

\begin{table}[t]
\centering
\resizebox{\linewidth}{!}{%
\begin{tabular}{@{}ccccc@{}}
\toprule
Finetuning Strategy & AUC@3$\uparrow$ & AUC@5$\uparrow$ & AUC@10$\uparrow$ & \multicolumn{1}{c}{AUC@30$\uparrow$} \\ \midrule
Train from Scratch                             &  5.10&  7.62&  25.53& 58.34 \\
Finetune from VGGT               &  \cellcolor[HTML]{EFEFEF}\textbf{28.18} &  \cellcolor[HTML]{EFEFEF}\textbf{45.51} &  \cellcolor[HTML]{EFEFEF}\textbf{65.31} & \cellcolor[HTML]{EFEFEF}\textbf{84.30}                     \\ \midrule
Freeze PTv3  &  25.00&  41.24&  61.32&   81.73                   \\ 
Finetune PTv3 & \cellcolor[HTML]{EFEFEF}\textbf{28.18} &  \cellcolor[HTML]{EFEFEF}\textbf{45.51} &  \cellcolor[HTML]{EFEFEF}\textbf{65.31} & \cellcolor[HTML]{EFEFEF}\textbf{84.30}                     \\
\bottomrule
\end{tabular}}
\vspace{-5pt}
\caption{\textbf{Effect of finetuning strategy.} Comparison of performance using different pretrained weight strategies.}
\vspace{-5pt}
\label{tab:ablate finetune}
\end{table}

\begin{table}[t]
\centering
\resizebox{\linewidth}{!}{%
\begin{tabular}{@{}lcccc@{}}
\toprule
\multicolumn{1}{c}{Geometry Injection Strategy} & AUC@3$\uparrow$ & AUC@5$\uparrow$ & AUC@10$\uparrow$ & \multicolumn{1}{c}{AUC@30$\uparrow$} \\ \midrule
VecSet + Raw Format              & 19.29&  34.72&  55.39&  78.56 \\
VecSet (w/ Texture) + Raw Format              & 22.85&  38.48&  57.42&  78.51 \\
PTv3 + Raw Format               &23.88&  40.22&  60.53&  79.52                      \\
PTv3 + View Format (Direct Add)  &  2.419&  4.996&  13.92& 41.81                     \\
PTv3 + View Format (CrossAttn)  & \cellcolor[HTML]{EFEFEF}\textbf{28.18} &  \cellcolor[HTML]{EFEFEF}\textbf{45.51} &  \cellcolor[HTML]{EFEFEF}\textbf{65.31} & \cellcolor[HTML]{EFEFEF}\textbf{84.30}                     \\
\bottomrule
\end{tabular}}
\vspace{-5pt}
\caption{\textbf{Comparison of geometry injection strategies.} We evaluate different geometry representations (VecSet~\cite{zhang20233dshape2vecset} and PTv3~\cite{wu2024ptv3}), their formats, and injection methods.}
\label{tab:ablate geometry injection}
\vspace{-15pt}
\end{table}

\vspace{-5pt}
\section{Limitations}
\label{sec: Limitations}

Although \awesomepose demonstrates strong generalization across diverse scenarios, it still assumes that the object remains static between the query images and the reference model. When the object undergoes non-rigid or articulated motion, the estimated pose may become inaccurate. A potential direction to address this limitation is to incorporate deformable component modeling into the framework, allowing the network to disentangle rigid motion from local shape deformation. Furthermore, since our network is trained primarily on solid, opaque objects, it struggles when applied to transparent or reflective objects. Such objects reveal background semantics or reflections that can mislead the model’s geometric reasoning and disrupt correspondence understanding. This limitation may be mitigated by introducing specialized preprocessing steps (\eg, background suppression or reflection removal) or by extending training data to include these challenging material types.

\section{Conclusion}
\label{sec: Conclusion}
In this work, we present a geometry-aware multi-view framework for unseen object pose estimation. Built upon recent multi-view foundation model architectures, the proposed approach incorporates object geometry through explicit point-based representations and learned geometry features projected into view-map form, facilitating more effective use of object model geometry information. Supported by our arranged large-scale synthetic dataset containing more than 190k objects under diverse conditions, the method demonstrates strong robustness and generalization across different scenarios. We believe this work represents an important step toward direct object pose estimation and toward integrating advances in 3D geometric reasoning within a unified framework.

\clearpage

\section*{Acknowledgements}
The research reported in this publication was supported by funding from King Abdullah University of Science and Technology (KAUST) – Center of Excellence for Generative AI, under award number 5940 and a gift from Google.

{
    \small
    \bibliographystyle{ieeenat_fullname}
    \bibliography{main}

@String(CVPR= {IEEE Conf. Comput. Vis. Pattern Recog.})

@String(ICCV= {Int. Conf. Comput. Vis.})

@String(ECCV= {Eur. Conf. Comput. Vis.})

@String(TOG= {ACM Trans. Graph.})

@String(VR   = {Vis. Res.})

@String(CVPR  = {CVPR})

@String(ICCV  = {ICCV})

@String(ECCV  = {ECCV})

@String(TOG   = {ACM TOG})

@inproceedings{okorn2021zephyr,
  title={Zephyr: Zero-shot pose hypothesis rating},
  author={Okorn, Brian and Gu, Qiao and Hebert, Martial and Held, David},
  booktitle={2021 IEEE International Conference on Robotics and Automation (ICRA)},
  pages={14141--14148},
  year={2021},
  organization={IEEE}
}

@inproceedings{sun2022ick,
  title={Ick-track: A category-level 6-dof pose tracker using inter-frame consistent keypoints for aerial manipulation},
  author={Sun, Jingtao and Wang, Yaonan and Feng, Mingtao and Wang, Danwei and Zhao, Jiawen and Stachniss, Cyrill and Chen, Xieyuanli},
  booktitle={2022 IEEE/RSJ International Conference on Intelligent Robots and Systems (IROS)},
  pages={1556--1563},
  year={2022},
  organization={IEEE}
}

@inproceedings{su2019deep,
  title={Deep multi-state object pose estimation for augmented reality assembly},
  author={Su, Yongzhi and Rambach, Jason and Minaskan, Nareg and Lesur, Paul and Pagani, Alain and Stricker, Didier},
  booktitle={2019 IEEE International Symposium on Mixed and Augmented Reality Adjunct (ISMAR-Adjunct)},
  pages={222--227},
  year={2019},
  organization={IEEE}
}

@inproceedings{li2024gbot,
  title={GBOT: Graph-based 3D object tracking for augmented reality-assisted assembly guidance},
  author={Li, Shiyu and Schieber, Hannah and Corell, Niklas and Egger, Bernhard and Kreimeier, Julian and Roth, Daniel},
  booktitle={2024 IEEE Conference Virtual Reality and 3D User Interfaces (VR)},
  pages={513--523},
  year={2024},
  organization={IEEE}
}

@article{hoque2023deep,
  title={Deep learning for 6D pose estimation of objects—A case study for autonomous driving},
  author={Hoque, Sabera and Xu, Shuxiang and Maiti, Ananda and Wei, Yuchen and Arafat, Md Yasir},
  journal={Expert Systems with Applications},
  volume={223},
  pages={119838},
  year={2023},
  publisher={Elsevier}
}

@inproceedings{wu20196d,
  title={6d-vnet: End-to-end 6-dof vehicle pose estimation from monocular rgb images},
  author={Wu, Di and Zhuang, Zhaoyong and Xiang, Canqun and Zou, Wenbin and Li, Xia},
  booktitle={Proceedings of the IEEE/CVF Conference on Computer Vision and Pattern Recognition Workshops},
  pages={0--0},
  year={2019}
}

@inproceedings{chen2025v2m4,
    title={V2M4: 4D Mesh Animation Reconstruction from a Single Monocular Video},
    author={Chen, Jianqi and Zhang, Biao and Tang, Xiangjun and Wonka, Peter},
    booktitle={Proceedings of the IEEE/CVF International Conference on Computer Vision (ICCV)},
    year={2025}
}

@article{yenphraphai2025shapegen4d,
  title={ShapeGen4D: Towards High Quality 4D Shape Generation from Videos},
  author={Yenphraphai, Jiraphon and Mirzaei, Ashkan and Chen, Jianqi and Zou, Jiaxu and Tulyakov, Sergey and Yeh, Raymond A and Wonka, Peter and Wang, Chaoyang},
  journal={arXiv preprint arXiv:2510.06208},
  year={2025}
}

@article{xiang2017posecnn,
  title={Posecnn: A convolutional neural network for 6d object pose estimation in cluttered scenes},
  author={Xiang, Yu and Schmidt, Tanner and Narayanan, Venkatraman and Fox, Dieter},
  journal={arXiv preprint arXiv:1711.00199},
  year={2017}
}

@inproceedings{peng2019pvnet,
  title={Pvnet: Pixel-wise voting network for 6dof pose estimation},
  author={Peng, Sida and Liu, Yuan and Huang, Qixing and Zhou, Xiaowei and Bao, Hujun},
  booktitle={Proceedings of the IEEE/CVF conference on computer vision and pattern recognition},
  pages={4561--4570},
  year={2019}
}

@inproceedings{li2019cdpn,
  title={Cdpn: Coordinates-based disentangled pose network for real-time rgb-based 6-dof object pose estimation},
  author={Li, Zhigang and Wang, Gu and Ji, Xiangyang},
  booktitle={Proceedings of the IEEE/CVF international conference on computer vision},
  pages={7678--7687},
  year={2019}
}

@inproceedings{rad2017bb8,
  title={Bb8: A scalable, accurate, robust to partial occlusion method for predicting the 3d poses of challenging objects without using depth},
  author={Rad, Mahdi and Lepetit, Vincent},
  booktitle={Proceedings of the IEEE international conference on computer vision},
  pages={3828--3836},
  year={2017}
}

@inproceedings{li2022dcl,
  title={Dcl-net: Deep correspondence learning network for 6d pose estimation},
  author={Li, Hongyang and Lin, Jiehong and Jia, Kui},
  booktitle={European Conference on Computer Vision},
  pages={369--385},
  year={2022},
  organization={Springer}
}

@inproceedings{sundermeyer2018implicit,
  title={Implicit 3d orientation learning for 6d object detection from rgb images},
  author={Sundermeyer, Martin and Marton, Zoltan-Csaba and Durner, Maximilian and Brucker, Manuel and Triebel, Rudolph},
  booktitle={Proceedings of the european conference on computer vision (ECCV)},
  pages={699--715},
  year={2018}
}

@inproceedings{wang2019densefusion,
  title={Densefusion: 6d object pose estimation by iterative dense fusion},
  author={Wang, Chen and Xu, Danfei and Zhu, Yuke and Mart{\'\i}n-Mart{\'\i}n, Roberto and Lu, Cewu and Fei-Fei, Li and Savarese, Silvio},
  booktitle={Proceedings of the IEEE/CVF conference on computer vision and pattern recognition},
  pages={3343--3352},
  year={2019}
}

@inproceedings{hu2020single,
  title={Single-stage 6d object pose estimation},
  author={Hu, Yinlin and Fua, Pascal and Wang, Wei and Salzmann, Mathieu},
  booktitle={Proceedings of the IEEE/CVF conference on computer vision and pattern recognition},
  pages={2930--2939},
  year={2020}
}

@article{zhang2023generative,
  title={Generative category-level object pose estimation via diffusion models},
  author={Zhang, Jiyao and Wu, Mingdong and Dong, Hao},
  journal={Advances in Neural Information Processing Systems},
  volume={36},
  pages={54627--54644},
  year={2023}
}

@inproceedings{lin2024instance,
  title={Instance-adaptive and geometric-aware keypoint learning for category-level 6d object pose estimation},
  author={Lin, Xiao and Yang, Wenfei and Gao, Yuan and Zhang, Tianzhu},
  booktitle={Proceedings of the IEEE/CVF Conference on Computer Vision and Pattern Recognition},
  pages={21040--21049},
  year={2024}
}

@inproceedings{tian2020shape,
  title={Shape prior deformation for categorical 6d object pose and size estimation},
  author={Tian, Meng and Ang Jr, Marcelo H and Lee, Gim Hee},
  booktitle={European Conference on Computer Vision},
  pages={530--546},
  year={2020},
  organization={Springer}
}

@inproceedings{wang2021category,
  title={Category-level 6d object pose estimation via cascaded relation and recurrent reconstruction networks},
  author={Wang, Jiaze and Chen, Kai and Dou, Qi},
  booktitle={2021 IEEE/RSJ International Conference on Intelligent Robots and Systems (IROS)},
  pages={4807--4814},
  year={2021},
  organization={IEEE}
}

@inproceedings{irshad2022centersnap,
  title={Centersnap: Single-shot multi-object 3d shape reconstruction and categorical 6d pose and size estimation},
  author={Irshad, Muhammad Zubair and Kollar, Thomas and Laskey, Michael and Stone, Kevin and Kira, Zsolt},
  booktitle={2022 International Conference on Robotics and Automation (ICRA)},
  pages={10632--10640},
  year={2022},
  organization={IEEE}
}

@inproceedings{lin2022sar,
  title={Sar-net: Shape alignment and recovery network for category-level 6d object pose and size estimation},
  author={Lin, Haitao and Liu, Zichang and Cheang, Chilam and Fu, Yanwei and Guo, Guodong and Xue, Xiangyang},
  booktitle={Proceedings of the IEEE/CVF conference on computer vision and pattern recognition},
  pages={6707--6717},
  year={2022}
}

@inproceedings{wang2019normalized,
  title={Normalized object coordinate space for category-level 6d object pose and size estimation},
  author={Wang, He and Sridhar, Srinath and Huang, Jingwei and Valentin, Julien and Song, Shuran and Guibas, Leonidas J},
  booktitle={Proceedings of the IEEE/CVF conference on computer vision and pattern recognition},
  pages={2642--2651},
  year={2019}
}

@inproceedings{fan2022object,
  title={Object level depth reconstruction for category level 6d object pose estimation from monocular rgb image},
  author={Fan, Zhaoxin and Song, Zhenbo and Xu, Jian and Wang, Zhicheng and Wu, Kejian and Liu, Hongyan and He, Jun},
  booktitle={European Conference on Computer Vision},
  pages={220--236},
  year={2022},
  organization={Springer}
}

@inproceedings{chen2021fs,
  title={Fs-net: Fast shape-based network for category-level 6d object pose estimation with decoupled rotation mechanism},
  author={Chen, Wei and Jia, Xi and Chang, Hyung Jin and Duan, Jinming and Shen, Linlin and Leonardis, Ales},
  booktitle={Proceedings of the IEEE/CVF conference on computer vision and pattern recognition},
  pages={1581--1590},
  year={2021}
}

@inproceedings{wang20196-pack,
  title={6-PACK: Category-level 6D Pose Tracker with Anchor-Based Keypoints},
  author={Wang, Chen and Mart{\'\i}n-Mart{\'\i}n, Roberto and Xu, Danfei and Lv, Jun and Lu, Cewu and Fei-Fei, Li and Savarese, Silvio and Zhu, Yuke},
  booktitle={International Conference on Robotics and Automation (ICRA)},
  year={2020}
}

@inproceedings{zhao2023learning,
  title={Learning symmetry-aware geometry correspondences for 6d object pose estimation},
  author={Zhao, Heng and Wei, Shenxing and Shi, Dahu and Tan, Wenming and Li, Zheyang and Ren, Ye and Wei, Xing and Yang, Yi and Pu, Shiliang},
  booktitle={Proceedings of the IEEE/CVF International Conference on Computer Vision},
  pages={14045--14054},
  year={2023}
}

@inproceedings{nguyen2024gigapose,
  title={Gigapose: Fast and robust novel object pose estimation via one correspondence},
  author={Nguyen, Van Nguyen and Groueix, Thibault and Salzmann, Mathieu and Lepetit, Vincent},
  booktitle={Proceedings of the IEEE/CVF Conference on Computer Vision and Pattern Recognition},
  pages={9903--9913},
  year={2024}
}

@inproceedings{nguyen2024nope,
  title={Nope: Novel object pose estimation from a single image},
  author={Nguyen, Van Nguyen and Groueix, Thibault and Ponimatkin, Georgy and Hu, Yinlin and Marlet, Renaud and Salzmann, Mathieu and Lepetit, Vincent},
  booktitle={Proceedings of the IEEE/CVF Conference on Computer Vision and Pattern Recognition},
  pages={17923--17932},
  year={2024}
}

@inproceedings{deng2025pos3r,
  title={Pos3R: 6D Pose Estimation for Unseen Objects Made Easy},
  author={Deng, Weijian and Campbell, Dylan and Sun, Chunyi and Zhang, Jiahao and Kanitkar, Shubham and Shaffer, Matt E and Gould, Stephen},
  booktitle={Proceedings of the Computer Vision and Pattern Recognition Conference},
  pages={16818--16828},
  year={2025}
}

@inproceedings{ausserlechner2024zs6d,
  title={Zs6d: Zero-shot 6d object pose estimation using vision transformers},
  author={Ausserlechner, Philipp and Haberger, David and Thalhammer, Stefan and Weibel, Jean-Baptiste and Vincze, Markus},
  booktitle={2024 IEEE International Conference on Robotics and Automation (ICRA)},
  pages={463--469},
  year={2024},
  organization={IEEE}
}

@article{chen2023zeropose,
  title={ZeroPose: CAD-model-based zero-shot pose estimation},
  author={Chen, Jianqiu and Sun, Mingshan and Bao, Tianpeng and Zhao, Rui and Wu, Liwei and He, Zhenyu},
  journal={arXiv e-prints},
  pages={arXiv--2305},
  year={2023}
}

@inproceedings{lin2024sam,
  title={Sam-6d: Segment anything model meets zero-shot 6d object pose estimation},
  author={Lin, Jiehong and Liu, Lihua and Lu, Dekun and Jia, Kui},
  booktitle={Proceedings of the IEEE/CVF Conference on Computer Vision and Pattern Recognition},
  pages={27906--27916},
  year={2024}
}

@inproceedings{huang2024matchu,
  title={Matchu: Matching unseen objects for 6d pose estimation from rgb-d images},
  author={Huang, Junwen and Yu, Hao and Yu, Kuan-Ting and Navab, Nassir and Ilic, Slobodan and Busam, Benjamin},
  booktitle={Proceedings of the IEEE/CVF Conference on Computer Vision and Pattern Recognition},
  pages={10095--10105},
  year={2024}
}

@article{geng2025one,
  title={One View, Many Worlds: Single-Image to 3D Object Meets Generative Domain Randomization for One-Shot 6D Pose Estimation},
  author={Geng, Zheng and Wang, Nan and Xu, Shaocong and Ye, Chongjie and Li, Bohan and Chen, Zhaoxi and Peng, Sida and Zhao, Hao},
  journal={arXiv preprint arXiv:2509.07978},
  year={2025}
}

@inproceedings{caraffa2024freeze,
  title={Freeze: Training-free zero-shot 6d pose estimation with geometric and vision foundation models},
  author={Caraffa, Andrea and Boscaini, Davide and Hamza, Amir and Poiesi, Fabio},
  booktitle={European Conference on Computer Vision},
  pages={414--431},
  year={2024},
  organization={Springer}
}

@inproceedings{ornek2024foundpose,
  title={Foundpose: Unseen object pose estimation with foundation features},
  author={{\"O}rnek, Evin P{\i}nar and Labb{\'e}, Yann and Tekin, Bugra and Ma, Lingni and Keskin, Cem and Forster, Christian and Hodan, Tomas},
  booktitle={European Conference on Computer Vision},
  pages={163--182},
  year={2024},
  organization={Springer}
}

@inproceedings{moon2024genflow,
  title={Genflow: Generalizable recurrent flow for 6d pose refinement of novel objects},
  author={Moon, Sungphill and Son, Hyeontae and Hur, Dongcheol and Kim, Sangwook},
  booktitle={Proceedings of the IEEE/CVF Conference on Computer Vision and Pattern Recognition},
  pages={10039--10049},
  year={2024}
}

@inproceedings{wen2024foundationpose,
  title={Foundationpose: Unified 6d pose estimation and tracking of novel objects},
  author={Wen, Bowen and Yang, Wei and Kautz, Jan and Birchfield, Stan},
  booktitle={Proceedings of the IEEE/CVF Conference on Computer Vision and Pattern Recognition},
  pages={17868--17879},
  year={2024}
}

@inproceedings{wang2025vggt,
  title={Vggt: Visual geometry grounded transformer},
  author={Wang, Jianyuan and Chen, Minghao and Karaev, Nikita and Vedaldi, Andrea and Rupprecht, Christian and Novotny, David},
  booktitle={Proceedings of the Computer Vision and Pattern Recognition Conference},
  pages={5294--5306},
  year={2025}
}

@article{wang2025pi,
  title={\ensuremath{\pi^3}: Scalable Permutation-Equivariant Visual Geometry Learning},
  author={Wang, Yifan and Zhou, Jianjun and Zhu, Haoyi and Chang, Wenzheng and Zhou, Yang and Li, Zizun and Chen, Junyi and Pang, Jiangmiao and Shen, Chunhua and He, Tong},
  journal={arXiv e-prints},
  pages={arXiv--2507},
  year={2025}
}

@inproceedings{leroy2024grounding,
  title={Grounding image matching in 3d with mast3r},
  author={Leroy, Vincent and Cabon, Yohann and Revaud, J{\'e}r{\^o}me},
  booktitle={European Conference on Computer Vision},
  pages={71--91},
  year={2024},
  organization={Springer}
}

@misc{oquab2023dinov2,
  title={DINOv2: Learning Robust Visual Features without Supervision},
  author={Oquab, Maxime and Darcet, Timothée and Moutakanni, Theo and Vo, Huy V. and Szafraniec, Marc and Khalidov, Vasil and Fernandez, Pierre and Haziza, Daniel and Massa, Francisco and El-Nouby, Alaaeldin and Howes, Russell and Huang, Po-Yao and Xu, Hu and Sharma, Vasu and Li, Shang-Wen and Galuba, Wojciech and Rabbat, Mike and Assran, Mido and Ballas, Nicolas and Synnaeve, Gabriel and Misra, Ishan and Jegou, Herve and Mairal, Julien and Labatut, Patrick and Joulin, Armand and Bojanowski, Piotr},
  journal={arXiv:2304.07193},
  year={2023}
}

@article{Laine2020diffrast,
  title   = {Modular Primitives for High-Performance Differentiable Rendering},
  author  = {Samuli Laine and Janne Hellsten and Tero Karras and Yeongho Seol and Jaakko Lehtinen and Timo Aila},
  journal = {ACM Transactions on Graphics},
  year    = {2020},
  volume  = {39},
  number  = {6}
}

@inproceedings{wu2024ptv3,
    title={Point Transformer V3: Simpler, Faster, Stronger},
    author={Wu, Xiaoyang and Jiang, Li and Wang, Peng-Shuai and Liu, Zhijian and Liu, Xihui and Qiao, Yu and Ouyang, Wanli and He, Tong and Zhao, Hengshuang},
    booktitle={CVPR},
    year={2024}
}

@article{deitke2023objaverse,
  title={Objaverse-xl: A universe of 10m+ 3d objects},
  author={Deitke, Matt and Liu, Ruoshi and Wallingford, Matthew and Ngo, Huong and Michel, Oscar and Kusupati, Aditya and Fan, Alan and Laforte, Christian and Voleti, Vikram and Gadre, Samir Yitzhak and others},
  journal={Advances in Neural Information Processing Systems},
  volume={36},
  pages={35799--35813},
  year={2023}
}

@inproceedings{collins2022abo,
  title={Abo: Dataset and benchmarks for real-world 3d object understanding},
  author={Collins, Jasmine and Goel, Shubham and Deng, Kenan and Luthra, Achleshwar and Xu, Leon and Gundogdu, Erhan and Zhang, Xi and Vicente, Tomas F Yago and Dideriksen, Thomas and Arora, Himanshu and others},
  booktitle={Proceedings of the IEEE/CVF conference on computer vision and pattern recognition},
  pages={21126--21136},
  year={2022}
}

@article{fu20213d,
  title={3d-future: 3d furniture shape with texture},
  author={Fu, Huan and Jia, Rongfei and Gao, Lin and Gong, Mingming and Zhao, Binqiang and Maybank, Steve and Tao, Dacheng},
  journal={International Journal of Computer Vision},
  volume={129},
  number={12},
  pages={3313--3337},
  year={2021},
  publisher={Springer}
}

@inproceedings{khanna2024habitat,
  title={Habitat synthetic scenes dataset (hssd-200): An analysis of 3d scene scale and realism tradeoffs for objectgoal navigation},
  author={Khanna, Mukul and Mao, Yongsen and Jiang, Hanxiao and Haresh, Sanjay and Shacklett, Brennan and Batra, Dhruv and Clegg, Alexander and Undersander, Eric and Chang, Angel X and Savva, Manolis},
  booktitle={Proceedings of the IEEE/CVF Conference on Computer Vision and Pattern Recognition},
  pages={16384--16393},
  year={2024}
}

@inproceedings{stojanov2021using,
  title={Using shape to categorize: Low-shot learning with an explicit shape bias},
  author={Stojanov, Stefan and Thai, Anh and Rehg, James M},
  booktitle={Proceedings of the IEEE/CVF conference on computer vision and pattern recognition},
  pages={1798--1808},
  year={2021}
}

@inproceedings{xiang2025structured,
  title={Structured 3d latents for scalable and versatile 3d generation},
  author={Xiang, Jianfeng and Lv, Zelong and Xu, Sicheng and Deng, Yu and Wang, Ruicheng and Zhang, Bowen and Chen, Dong and Tong, Xin and Yang, Jiaolong},
  booktitle={Proceedings of the Computer Vision and Pattern Recognition Conference},
  pages={21469--21480},
  year={2025}
}

@misc{polyheaven,
  author       = {Poly Haven},
  title        = {Poly Haven: Public 3D Asset Library},
  howpublished = {\url{https://polyhaven.com/}},
  note         = {Accessed: 2025-11-03}
}

@article{chen2025zero,
  title={Zero-shot image harmonization with generative model prior},
  author={Chen, Jianqi and Zhang, Yilan and Zou, Zhengxia and Chen, Keyan and Shi, Zhenwei},
  journal={IEEE Transactions on Multimedia},
  year={2025},
  publisher={IEEE}
}

@inproceedings{mokady2023null,
  title={Null-text inversion for editing real images using guided diffusion models},
  author={Mokady, Ron and Hertz, Amir and Aberman, Kfir and Pritch, Yael and Cohen-Or, Daniel},
  booktitle={Proceedings of the IEEE/CVF conference on computer vision and pattern recognition},
  pages={6038--6047},
  year={2023}
}

@article{song2020denoising,
  title={Denoising diffusion implicit models},
  author={Song, Jiaming and Meng, Chenlin and Ermon, Stefano},
  journal={arXiv preprint arXiv:2010.02502},
  year={2020}
}

@misc{flux1_canny_dev,
  author       = {Black Forest Labs},
  title        = {FLUX.1-Canny-dev},
  howpublished = {\url{https://huggingface.co/black-forest-labs/FLUX.1-Canny-dev}},
  note         = {Accessed: 2025-11-03},
  year         = {2024}
}

@article{jiang2025rayzer,
  title={RayZer: A Self-supervised Large View Synthesis Model},
  author={Jiang, Hanwen and Tan, Hao and Wang, Peng and Jin, Haian and Zhao, Yue and Bi, Sai and Zhang, Kai and Luan, Fujun and Sunkavalli, Kalyan and Huang, Qixing and others},
  journal={arXiv preprint arXiv:2505.00702},
  year={2025}
}

@inproceedings{hodavn2020bop,
  title={BOP challenge 2020 on 6D object localization},
  author={Hoda{\v{n}}, Tom{\'a}{\v{s}} and Sundermeyer, Martin and Drost, Bertram and Labb{\'e}, Yann and Brachmann, Eric and Michel, Frank and Rother, Carsten and Matas, Ji{\v{r}}{\'\i}},
  booktitle={European Conference on Computer Vision},
  pages={577--594},
  year={2020},
  organization={Springer}
}

@article{hodan2017tless,
  title={{T-LESS}: An {RGB-D} Dataset for {6D} Pose Estimation of Texture-less Objects},
  author={Hoda{\v{n}}, Tom{\'a}{\v{s}} and Haluza, Pavel and Obdr{\v{z}}{\'a}lek, {\v{S}}t{\v{e}}p{\'a}n and Matas, Ji{\v{r}}{\'\i} and Lourakis, Manolis and Zabulis, Xenophon},
  journal={WACV},
  year={2017}
}

@article{xiang2018posecnn,
    Author = {Xiang, Yu and Schmidt, Tanner and Narayanan, Venkatraman and , Dieter},
    Title = {PoseCNN: A Convolutional Neural Network for 6D Object Pose Estimation in Cluttered Scenes},
    Journal ={Robotics: Science and Systems},
    Year = {2018}
}

@InProceedings{hodan2018bop,
author = {Hodan, Tomas and Michel, Frank and Brachmann, Eric and Kehl, Wadim and GlentBuch, Anders and Kraft, Dirk and Drost, Bertram and Vidal, Joel and Ihrke, Stephan and Zabulis, Xenophon and Sahin, Caner and Manhardt, Fabian and Tombari, Federico and Kim, Tae-Kyun and Matas, Jiri and Rother, Carsten},
title = {BOP: Benchmark for 6D Object Pose Estimation},
booktitle = {ECCV},
year = {2018}
}

@inproceedings{doumanoglou2016recovering,
  title={Recovering 6D object pose and predicting next-best-view in the crowd},
  author={Doumanoglou, Andreas and Kouskouridas, Rigas and Malassiotis, Sotiris and Kim, Tae-Kyun},
  booktitle={CVPR},
  year={2016}
}

@inproceedings{brachmann2014learning,
  title={Learning 6d object pose estimation using 3d object coordinates},
  author={Brachmann, Eric and Krull, Alexander and Michel, Frank and Gumhold, Stefan and Shotton, Jamie and Rother, Carsten},
  booktitle={ECCV},
  year={2014},
}

@inproceedings{osop,
  title={Osop: A multi-stage one shot object pose estimation framework},
  author={Shugurov, Ivan and Li, Fu and Busam, Benjamin and Ilic, Slobodan},
  booktitle={CVPR},
  year={2022}
}

@inproceedings{megapose,
    title     = {MegaPose: 6D Pose Estimation of Novel Objects via Render \& Compare},
    author    = {Labb\'e, Yann and Manuelli, Lucas and Mousavian, Arsalan and Tyree, Stephen and Birchfield, Stan and Tremblay, Jonathan and Carpentier, Justin and Aubry, Mathieu and Fox, Dieter and Sivic, Josef},
    booktitle = {CoRL},
    year = {2022},
  }

@inproceedings{huang2025raypose,
  title={RayPose: Ray Bundling Diffusion for Template Views in Unseen 6D Object Pose Estimation},
  author={Huang, Junwen and Vutukur, Shishir Reddy and Yu, Peter KT and Navab, Nassir and Ilic, Slobodan and Busam, Benjamin},
  booktitle={Proceedings of the IEEE/CVF International Conference on Computer Vision},
  pages={9102--9112},
  year={2025}
}

@inproceedings{wu2025sonata,
  title={Sonata: Self-supervised learning of reliable point representations},
  author={Wu, Xiaoyang and DeTone, Daniel and Frost, Duncan and Shen, Tianwei and Xie, Chris and Yang, Nan and Engel, Jakob and Newcombe, Richard and Zhao, Hengshuang and Straub, Julian},
  booktitle={Proceedings of the Computer Vision and Pattern Recognition Conference},
  pages={22193--22204},
  year={2025}
}

@article{li2025step1x,
  title={Step1x-3d: Towards high-fidelity and controllable generation of textured 3d assets},
  author={Li, Weiyu and Zhang, Xuanyang and Sun, Zheng and Qi, Di and Li, Hao and Cheng, Wei and Cai, Weiwei and Wu, Shihao and Liu, Jiarui and Wang, Zihao and others},
  journal={arXiv preprint arXiv:2505.07747},
  year={2025}
}

@article{zhang20233dshape2vecset,
  title={{3DShape2VecSet}: A 3d shape representation for neural fields and generative diffusion models},
  author={Zhang, Biao and Tang, Jiapeng and Niessner, Matthias and Wonka, Peter},
  journal={ACM Transactions On Graphics (TOG)},
  volume={42},
  number={4},
  pages={1--16},
  year={2023},
  publisher={ACM New York, NY, USA}
}

@inproceedings{wang2024dust3r,
  title={Dust3r: Geometric 3d vision made easy},
  author={Wang, Shuzhe and Leroy, Vincent and Cabon, Yohann and Chidlovskii, Boris and Revaud, Jerome},
  booktitle={Proceedings of the IEEE/CVF Conference on Computer Vision and Pattern Recognition},
  pages={20697--20709},
  year={2024}
}

@article{crandall2012sfm,
  title={SfM with MRFs: Discrete-continuous optimization for large-scale structure from motion},
  author={Crandall, David J and Owens, Andrew and Snavely, Noah and Huttenlocher, Daniel P},
  journal={IEEE transactions on pattern analysis and machine intelligence},
  volume={35},
  number={12},
  pages={2841--2853},
  year={2012},
  publisher={IEEE}
}

@inproceedings{schonberger2016structure,
  title={Structure-from-motion revisited},
  author={Schonberger, Johannes L and Frahm, Jan-Michael},
  booktitle={Proceedings of the IEEE conference on computer vision and pattern recognition},
  pages={4104--4113},
  year={2016}
}

@inproceedings{cui2017hsfm,
  title={HSfM: Hybrid structure-from-motion},
  author={Cui, Hainan and Gao, Xiang and Shen, Shuhan and Hu, Zhanyi},
  booktitle={Proceedings of the IEEE conference on computer vision and pattern recognition},
  pages={1212--1221},
  year={2017}
}

@inproceedings{goesele2006multi,
  title={Multi-view stereo revisited},
  author={Goesele, Michael and Curless, Brian and Seitz, Steven M},
  booktitle={2006 IEEE Computer Society Conference on Computer Vision and Pattern Recognition (CVPR'06)},
  volume={2},
  pages={2402--2409},
  year={2006},
  organization={IEEE}
}

@article{liu2024deep,
  title={Deep learning-based object pose estimation: A comprehensive survey},
  author={Liu, Jian and Sun, Wei and Yang, Hui and Zeng, Zhiwen and Liu, Chongpei and Zheng, Jin and Liu, Xingyu and Rahmani, Hossein and Sebe, Nicu and Mian, Ajmal},
  journal={arXiv preprint arXiv:2405.07801},
  year={2024}
}

@inproceedings{yu2019free,
  title={Free-form image inpainting with gated convolution},
  author={Yu, Jiahui and Lin, Zhe and Yang, Jimei and Shen, Xiaohui and Lu, Xin and Huang, Thomas S},
  booktitle={Proceedings of the IEEE/CVF international conference on computer vision},
  pages={4471--4480},
  year={2019}
}

@InProceedings{Nguyen_2023_ICCV,
    author    = {Nguyen, Van Nguyen and Groueix, Thibault and Ponimatkin, Georgy and Lepetit, Vincent and Hodan, Tomas},
    title     = {CNOS: A Strong Baseline for CAD-Based Novel Object Segmentation},
    booktitle = {Proceedings of the IEEE/CVF International Conference on Computer Vision (ICCV) Workshops},
    month     = {October},
    year      = {2023},
    pages     = {2134-2140}
}

@software{blender,
  author = {{Blender Foundation}},
  title  = {Blender},
  year   = {2024},
  url    = {https://www.blender.org/},
}
}

\clearpage

\appendix

\section{Overview}

In this supplementary material, we first examine the fragility of multi-view foundation models under appearance-inconsistent inputs in Appendix~\ref{app: fragile}. We then provide additional details about our constructed dataset in Appendix~\ref{app: dataset}. Appendix~\ref{app: implement} presents the implementation details of our network architecture, training procedure, inference pipeline, and evaluation setup. In Appendix~\ref{app: more quantitative}, we provide additional quantitative comparisons with existing methods, including runtime and memory costs, the number of reference images, and performance after applying refinement networks, along with ablation studies of our dataset. Finally, we include additional qualitative comparison results in Appendix~\ref{app: visual}.

\section{Fragility of Multi-View Foundation Models to Appearance-Inconsistent Inputs}
\label{app: fragile}

\begin{figure}[!t]
  \centering
  \includegraphics[width=\linewidth]{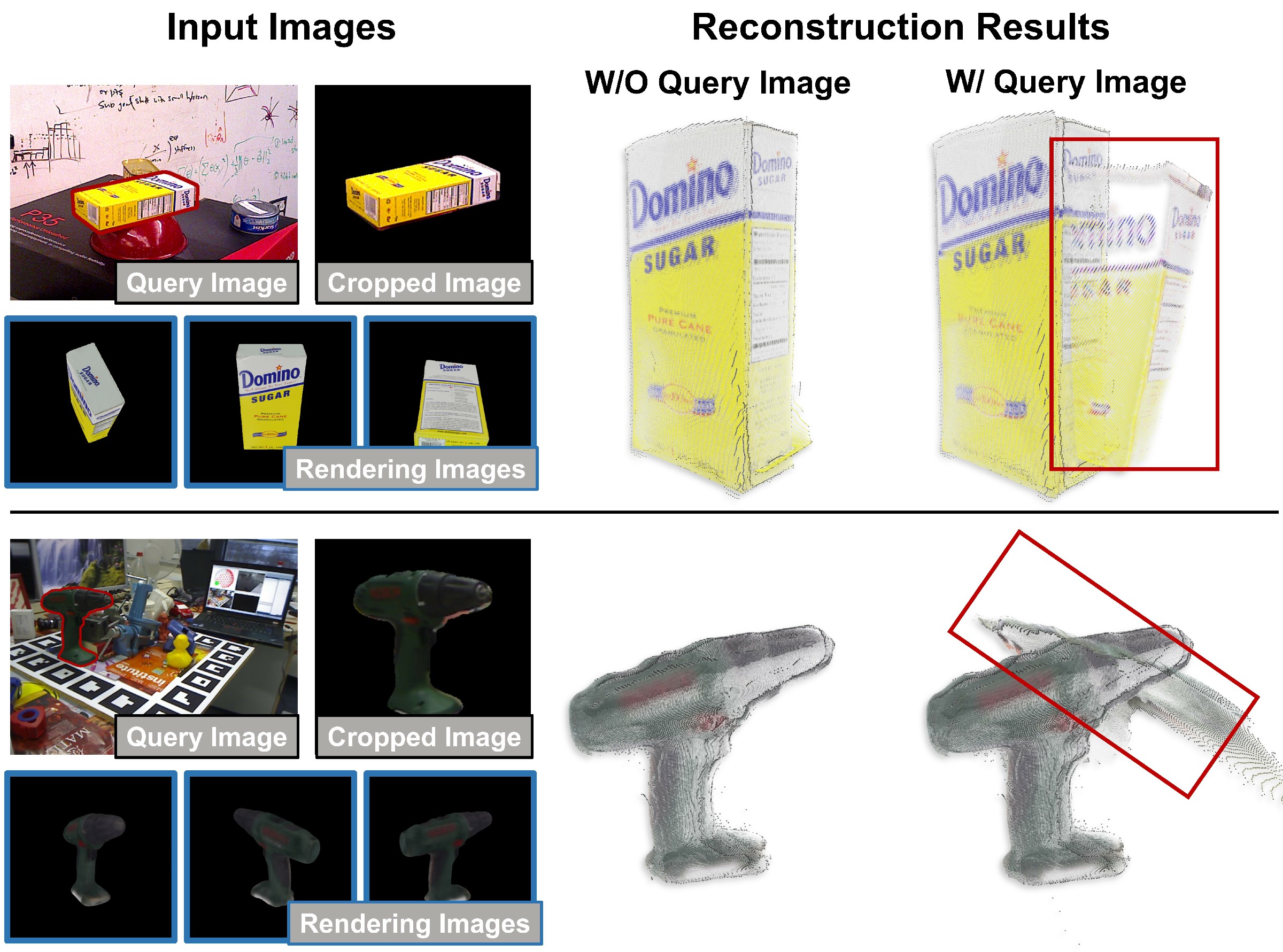}
  \caption{\textbf{Analysis of the Fragility of Multi-View Foundation Models to Appearance-Inconsistent Inputs.}
The input set consists of several rendered images along with a cropped region from the query image. When only rendered images are provided, VGGT~\cite{wang2025vggt} predicts consistent poses across views, as evidenced by the clean and coherent reconstructed point cloud without outliers. These point clouds are obtained by back-projecting image pixels into 3D using the predicted camera poses.
However, once the cropped query image is included, VGGT fails to maintain consistency: the predicted pose for the query image becomes inaccurate. The red bounding box highlights the resulting deviations in the reconstructed point cloud, demonstrating the sensitivity of the model to appearance-inconsistent inputs.}
  \vspace{-10pt}
  \label{fig:analysis}
\end{figure}

In Fig.~\ref{fig:analysis}, we illustrate the fragility of multi-view foundation models when confronted with appearance-inconsistent inputs. The cropped image originates from the real-world query image, whereas the rendering views are generated from the CAD model in a virtual space. Due to differences in lighting, material properties, and other photometric effects, the object’s appearance in the query image differs subtly from that in the rendering images.

Since multi-view foundation models such as VGGT~\cite{wang2025vggt} are not trained to consider this appearance gap between real and synthetic domains, they often produce highly inaccurate predictions under such mismatched conditions. In Fig.~\ref{fig:analysis}, we show the reconstruction results using only the rendering images and using both the rendering images and the query image. As highlighted by the red bounding boxes, the point cloud projected from the query image using the predicted camera pose deviates substantially from the points projected from the rendering images. These inconsistencies appear as large outliers, indicating that the model fails to correctly estimate the camera pose and is therefore sensitive to appearance-inconsistent multi-frame inputs.

This observation also highlights the necessity and practical value of our appearance-editing data type in the large-scale dataset (see Sec.~\ref{subsec: image data} in the main paper), which helps mitigate such synthetic-to-real appearance gaps.

\begin{figure*}[!t]
  \centering
  \includegraphics[width=\linewidth]{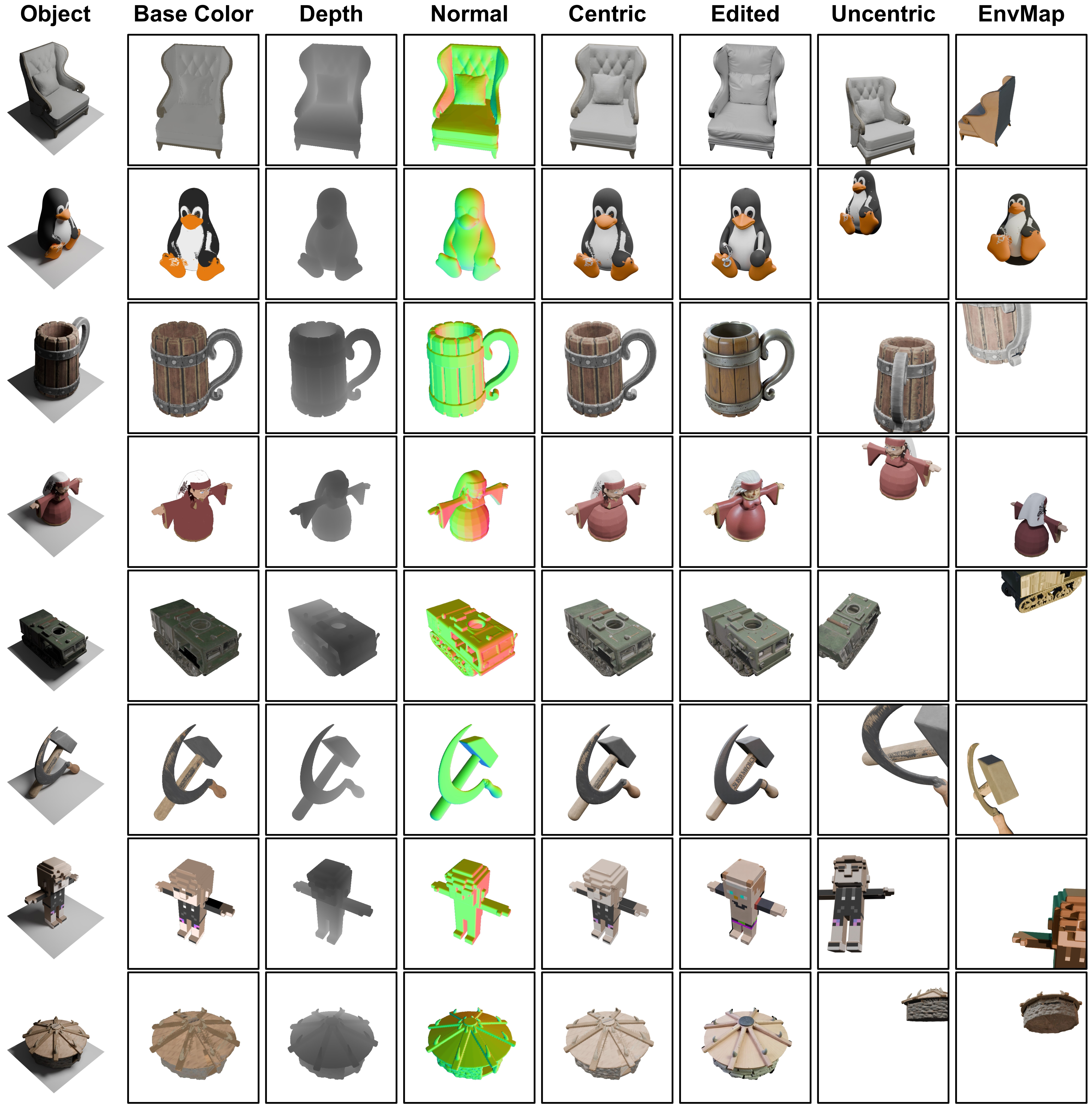}
  \caption{\textbf{Examples from the constructed object pose estimation dataset.} From left to right: object mesh after texture rebaking, basecolor rendering image, depth map, normal map, and four types of rendered images used as query inputs during training (see Sec.~\ref{subsec: image data} in the main paper for details).}
  \label{fig:dataset2}
\end{figure*}

\section{Additional Dataset Construction Details}
\label{app: dataset}

Given access to the object model $\mathcal{M}$, we synthesize additional geometry-related data beyond the texture-rendered image $I_{i}$ under each manually defined camera pose $T_{i}$. Specifically, we first render a depth map $D_i$ at pose $T_{i}$ using a graphics renderer such as Blender~\cite{blender} or Nvdiffrast~\cite{Laine2020diffrast}. Since both the camera pose $T_{i}$ and the intrinsic matrix of the rendering camera are known, we can reconstruct the corresponding point map in world coordinates, denoted as $P_{i}$, through the standard camera projection model.

In addition to depth, we also render the normal map $O_i$ and the object mask $M_i$ using the graphics renderer. These geometry-aware data modalities satisfy the input requirements of the PTv3~\cite{wu2024ptv3} geometry representation network used in our method (see Appendix~\ref{app: network implement}).

Fig.~\ref{fig:dataset2} presents additional visual examples to illustrate the full composition of our dataset. We also compare the statistics of our dataset with existing datasets in Appendix~\ref{app: more quantitative}, along with an ablation study analyzing the contribution of each subset.

\section{Implementation Details}
\label{app: implement}

\subsection{Network Architecture}
\label{app: network implement}

\textbf{Input Data.} As illustrated in Fig.~\ref{fig:framework} of the main paper, the network input consists of two components: the query image $I_{\text{query}}$ and multi-view data rendered around the object model $\mathcal{M}$. The multi-view data comprises four modalities: camera parameters $\mathcal{T}$, RGB images $\mathcal{V}$, point maps $\mathcal{P}$, and geometric feature maps $\mathcal{F}$. These modalities are paired such that the elements at the same index correspond to the same viewpoint of the object model. Note that the point maps $\mathcal{P}$ and geometric feature maps $\mathcal{F}$ are not directly available; they are derived from depth maps and camera intrinsics, and from point maps, as described in Appendix~\ref{app: dataset} and detailed further below.

\paragraph{Geometry Feature Extractor.} As described in Sec.~\ref{sec:point-net} of the main paper, we employ a geometry representation network to extract a global 3D representation of the object. Specifically, we use PTv3~\cite{wu2024ptv3}. The original PTv3 takes as input a point cloud $\mathbb{R}^{S \times 9}$, where $S$ denotes the total number of points and each point encodes 3D coordinates, surface normals, and color features.

To allow rearrangement of the extracted global representation into the view-based format, we use point maps $\mathcal{P}$ instead of directly sampling points from the object mesh surface. Given that the camera poses $\mathcal{T}$ are carefully sampled to cover the entire object (see Appendix~\ref{app: training and inference implement}), the aggregation of point maps across all rendered views sufficiently represents the full object and serves as input to PTv3. Specifically, for each rendered view, we use the object mask to select valid pixels and construct the per-view point cloud:
\begin{equation}
    \text{PC}_{i} = \operatorname{cat}\!\big[P_{i}, O_{i}, I_{i}\big]_{M_{i}}
\end{equation}
Each point in $\text{PC}_{i}$ contains a 3D coordinate, a surface normal, and an RGB color. All per-view point clouds are then concatenated to form the final input point cloud: $\text{PC} = \bigcup_{i=1}^{N} \text{PC}_{i}$.

After encoding with PTv3, each point is embedded into a feature vector ($\mathbf{e}_i \in \mathbb{R}^C$). To redistribute these features back into a multi-view format, we restore their spatial positions using the mask indices that record each point’s source pixel coordinates $(u_i, v_i)$ in the corresponding view. The per-view feature map is reconstructed as:
\begin{equation}
{F}_{i}(u,v) =
\begin{cases}
\mathbf{e}_i, & \text{if } (u,v) = (u_i, v_i) \text{ for point } i,\\[3pt]
\mathbf{0}, & \text{otherwise.}
\end{cases}
\end{equation}

In practice, since each viewpoint contains a large number of pixels (over half a million), aggregating all views at full resolution would incur excessive computational and memory costs. To address this, we downsample the inputs $P_{i}$, $O_{i}$, and $I_{i}$ prior to feature extraction.

\paragraph{Multi-Modal Encoders.} The network employs four separate encoders to process each modality. For RGB images, we use DINOv2~\cite{oquab2023dinov2} as the image encoder, producing feature tokens $X$. Camera parameters are represented as a 9-dimensional vector, comprising 4 parameters for rotation (quaternion), 3 for translation, and 2 for the field of view. These are encoded into a single camera token $\mathbf{c}$ using a simple MLP. For the point maps and geometric feature maps, we apply a strided convolutional layer followed by flattening, producing patch tokens for each modality.

\paragraph{Fusion Transformer.} The tokens from all encoders are processed by a stack of 24 attention blocks, each consisting of both cross-attention and self-attention layers. In the cross-attention layers, the combination of image tokens and camera tokens $(X, \mathbf{c})$ serves as the query. Since the query image $I_{\text{query}}$ has no known camera extrinsics, a learnable token is assigned in its place. The point map tokens and geometric feature tokens are used as the key and value tokens in the cross-attention layers.

For the self-attention layers, we adopt the alternating intra-frame and inter-frame self-attention mechanism from VGGT~\cite{wang2025vggt}. Intra-frame self-attention computes relationships among tokens within the same viewpoint, while inter-frame self-attention captures correlations across tokens from different viewpoints, modeling multi-view interactions.

\paragraph{Pose Output Decoder.} From the output tokens of the Fusion Transformer, we select only those corresponding to the input camera tokens (for the query image, this corresponds to the learnable token). These tokens, one per viewpoint, are passed to a camera head whose architecture is inherited from VGGT~\cite{wang2025vggt}. The camera head consists of several self-attention layers followed by an MLP, which decodes each token into a 9-dimensional vector representing the camera extrinsics and intrinsics in the same format described above.

\subsection{Model Training, Inference, and Evaluation}
\label{app: training and inference implement}

\textbf{Training Details.} During training, we freeze the pretrained DINOv2~\cite{oquab2023dinov2} weights for image feature extraction. For PTv3, we use pretrained weights from Sonata~\cite{wu2025sonata} and set its learning rate to $5 \times 10^{-5}$. For the self-attention layers and the camera head, pretrained weights from VGGT~\cite{wang2025vggt} are used, with the same fine-tuning learning rate of $5 \times 10^{-5}$. All other layers use a learning rate of $2 \times 10^{-4}$, including the cross-attention layers and the encoders for camera, point maps, and geometric feature maps. The full network contains approximately 1.5B parameters and is trained for 100k iterations on 4 A100 GPUs, taking roughly 8 days.

To improve training efficiency, we adopt dynamic batch training, similar to VGGT~\cite{wang2025vggt}, where each object is represented by a randomly selected 11–24 multi-view frames. Ten of these camera poses are used as known views, sampled via farthest point sampling (FPS) from the 50 camera poses prepared for each object (see Sec.~\ref{subsec: image data} in the main paper) to ensure full coverage. For these known views, we use basecolor-rendered images (see Appendix~\ref{app: dataset}) to minimize the influence of environmental lighting. The remaining frames are used as unknown query images, randomly selected from the four types of rendered images (Sec.~\ref{subsec: image data} in the main paper). The maximum number of frames per batch is set to 48.

For data augmentation, in addition to standard transformations such as color jittering and grayscale conversion, we apply random 3D rotations to the input object models. Corresponding rotations are applied to normal maps, point maps, and camera extrinsics to maintain alignment. We also apply 2D image-space random rotations to the RGB inputs, with corresponding adjustments to normal maps, point maps, depth maps, mask maps, and extrinsics for each frame. To simulate practical occlusion scenarios, objects are randomly masked using ellipses, rectangles, and free-form masks~\cite{yu2019free}. To better model noise introduced by imperfect object segmentation in query images, we randomly overlay or underlay images of other objects onto the query image. Additionally, to mimic blur artifacts introduced during cropping and scaling, we apply JPEG compression and Gaussian blur to the images.

To stabilize training despite varying object sizes, all objects are normalized to fit within a bounding box of $[-B, B]$ along each axis, with $B=1.05$. The training objective is a camera pose regression loss, formulated as an L1 loss between predicted and ground-truth camera parameters. Both known and query views are supervised, but the loss from known views is weighted by 0.5 to encourage the network to focus on the more challenging query-view estimation. Unlike VGGT~\cite{wang2025vggt}, which predicts relative poses between frames, our method directly predicts absolute camera poses, allowing direct use of the geometry from the object model $\mathcal{M}$ defined in an absolute coordinate system.

\paragraph{Inference Details.} To remain consistent with training in a normalized object space, inference is performed on a normalized version of the object. The predicted pose is then rescaled to account for the normalization, producing the final object pose in the real-world scale. 

Specifically, given a query image $I_\text{query}$ and an object model $\mathcal{M}$, we first normalize $\mathcal{M}$ into the canonical bounding box used during training. Let $s$ denote the isotropic scaling factor that maps points from the original object to the normalized one. The network predicts the camera pose relative to the normalized object, denoted as $T^{\text{\text{Cam}}}_{\text{norm}}$. Since the desired output is the object-to-camera transformation, we invert this prediction: $T^{\text{norm}}_{\text{Cam}} = (T^{\text{\text{Cam}}}_{\text{norm}})^{-1}$.

To rescale this normalized extrinsic back to the true object scale, consider any 3D point $\mathbf{q}$ in the original object frame. Its normalized counterpart is $\mathbf{q}_{\text{norm}} = s\,\mathbf{q}$. Let the network-predicted camera pose with respect to the normalized object be $T^{\text{norm}}_{\text{Cam}}
= [\mathbf{R}_{\text{norm}},\,\mathbf{t}_{\text{norm}}]$. We denote perspective projection under camera intrinsics $K$ by (we slightly abuse notation here and let $\mathbf{q}$ denote the transformed 3D point in the camera frame):
\begin{equation}
\pi(\mathbf{q})
=
\mathrm{div}\!\left( K\mathbf{q} \right)
=
\begin{bmatrix}
\displaystyle \frac{(K\mathbf{q})_x}{(K\mathbf{q})_z} ,
\displaystyle \frac{(K\mathbf{q})_y}{(K\mathbf{q})_z}
\end{bmatrix}
\end{equation}
where $\mathrm{div}$ indicates division by the depth coordinate, and
$(\cdot)_x$, $(\cdot)_y$, $(\cdot)_z$ denote the horizontal, vertical, and depth components.

To ensure that the projected 2D pixel location of the object remains unchanged
before and after rescaling, we seek a rotation $\mathbf{R}$ and a translation
$\mathbf{t}$ such that:
\begin{equation}
    \pi\!\left(
\mathbf{R}_{\text{norm}}(s\,\mathbf{q}) + \mathbf{t}_{\text{norm}}
\right)
=
\pi\!\left(
\mathbf{R}\mathbf{q} + \mathbf{t}
\right)
\end{equation}

Since both sides use the same intrinsics and perspective projection is defined up to an arbitrary nonzero scale before depth division, this is equivalent to enforcing:
\begin{equation}
    \frac{
\mathbf{R}_{\text{norm}}(s\,\mathbf{q})+\mathbf{t}_{\text{norm}}
}{
\left[\mathbf{R}_{\text{norm}}(s\,\mathbf{q})+\mathbf{t}_{\text{norm}}\right]_z
}
=
\frac{
\mathbf{R}\mathbf{q}+\mathbf{t}
}{
\left[\mathbf{R}\mathbf{q}+\mathbf{t}\right]_z
}
\end{equation}

Dividing the numerator and denominator on the left-hand side by $s$ gives:
\begin{equation}
    \frac{
\mathbf{R}_{\text{norm}}\mathbf{q}
+ \mathbf{t}_{\text{norm}}/s
}{
\left[\mathbf{R}_{\text{norm}}\mathbf{q}
+ \mathbf{t}_{\text{norm}}/s\right]_z
}
=
\frac{
\mathbf{R}\mathbf{q}+\mathbf{t}
}{
\left[\mathbf{R}\mathbf{q}+\mathbf{t}\right]_z
}
\end{equation}

Therefore, the pose estimation for the real object is:
\begin{equation}
T_{\text{query}} =
[\mathbf{R},\, \mathbf{t}]
= [\mathbf{R}_{\text{norm}},\, \frac{\mathbf{t}_{\text{norm}}}{s}]
\end{equation}

\paragraph{Evaluation Details.} When evaluating our method on BOP benchmark datasets, we have access to the 3D object model, the query image, the camera intrinsics for the query image, and the segmentation (from CNOS~\cite{Nguyen_2023_ICCV}) of the target object. We first select 10 camera poses around the 3D object (ensuring full coverage via farthest point sampling), using intrinsics the same as the query image, to render multi-view frames. For the query image, since the object often occupies only a small region, we crop the object using the available segmentation map and resize it to the same resolution as the original query image. We then perform pose estimation following the procedure described in the Inference Details above.

Since the network outputs extrinsics relative to the output intrinsics, which differ from the target intrinsics, we convert the estimated pose $(\mathbf{R}_{\text{src}}, \mathbf{t}_{\text{src}})$ to the target intrinsic space using SVD. Specifically, we map the source projection matrix into the target intrinsic space via $K_{\text{dst}}^{-1}K_{\text{src}} [\mathbf{R}_{\text{src}} \,|\, \mathbf{t}_{\text{src}}]$, and extract the closest rotation and translation through SVD, yielding $(\mathbf{R}_{\text{dst}}, \mathbf{t}_{\text{dst}})$ whose projection under $K_{\text{dst}}$ approximates the original image.

However, since $K_{\text{dst}}^{-1} K_{\text{src}}$ contains non-uniform scalings and translations, the rotation matrix cannot be recovered accurately. Consequently, the SVD-based rotation--translation approximation introduces error, preventing recovery of the extrinsics. To further compensate for this, we leverage an IoU loss and a 2D Chamfer loss to compare the input segmentation map with the rendering mask generated using the SVD-converted extrinsics. The extrinsics are then updated through some iterations of gradient descent. Although this solution depends on the quality of the segmentation map, it generally produces reliable results in our evaluation.

\begin{figure*}[!t]
  \centering
  \includegraphics[width=\linewidth]{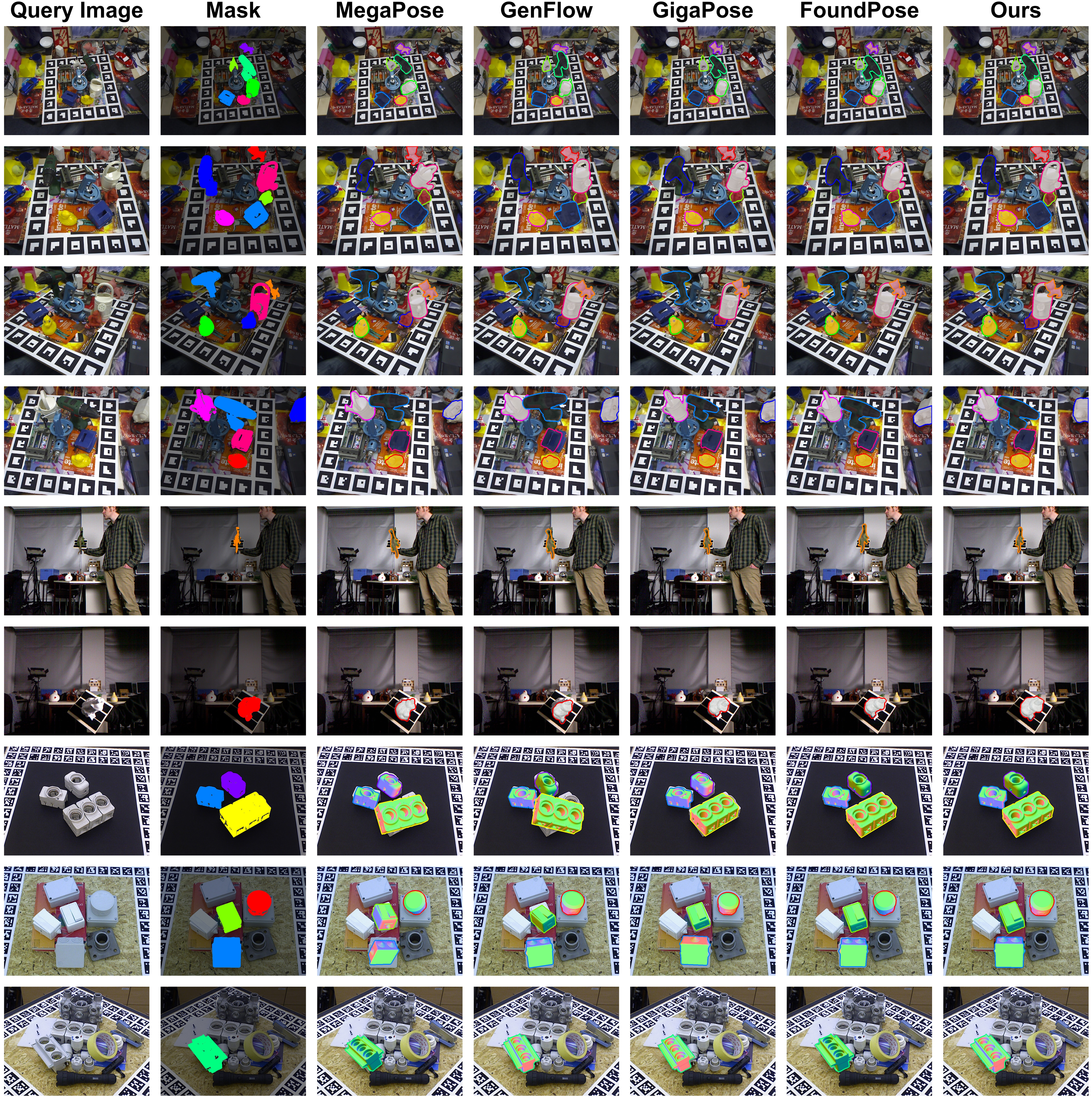}
  \caption{\textbf{Visual comparison with other methods.} From left to right: the query image for pose estimation, object masks with each object shown in a distinct color, and the projection results of different methods after applying the estimated poses to the 3D object models. The projected 3D models are outlined with borders matching the colors of their corresponding masks. The last three rows show samples from the T-LESS dataset~\cite{hodan2017tless}; since this dataset does not provide object textures, we display the rendered normal images of the objects for clearer visualization.}
  \label{fig:comparison2}
\end{figure*}

\section{Additional Quantitative Results}
\label{app: more quantitative}

\paragraph{Runtime, Memory Cost, and Number of Reference Views.}
Table~\ref{Tab: complexity} presents runtime and memory comparisons across different methods, while Table~\ref{Tab: reference image number} reports the number of reference views used by each method for Table~\ref{tab:compare} in the main paper. As shown, our method requires fewer reference views than competing approaches. Benefiting from this design, our method incurs negligible onboarding cost. MegaPose~\cite{megapose} also achieves low onboarding cost; however, it shifts the computational burden to the inference stage. In terms of memory consumption, our method requires more memory due to its larger network size, but still maintains competitive inference speed.

\begin{table}[t]
\renewcommand{\arraystretch}{1.3}
\resizebox{\columnwidth}{!}{\begin{tabular}{@{}c|ccc@{}}
\toprule
\multirow{2}{*}[-5pt]{Method} & \multicolumn{2}{c}{Time per Object Detection} & \multirow{2}{*}[-5pt]{Peak GPU Memory} \\ \cmidrule(lr){2-3}
 & \begin{tabular}[c]{@{}c@{}}Onboarding \\ (Offline, e.g., rendering)\end{tabular} & \begin{tabular}[c]{@{}c@{}}Pose Estimation \\ (Inference)\end{tabular} &  \\ \midrule
MegaPose & $\sim$0s & 1.47s & 3.33GB \\
GigaPose & 1.77s & 0.26s & 6.47GB \\
FoundPose & 1100.37s & 0.46s & 0.83GB \\
Ours & 0.30s & 0.67s & 9.49GB \\ \bottomrule
\end{tabular}
}
\caption{Runtime and peak GPU memory comparisons, measured per object detection.}\label{Tab: complexity}
\end{table}

\begin{table}[t]
\renewcommand{\arraystretch}{1.3}
\resizebox{\columnwidth}{!}{\begin{tabular}{@{}c|cccc@{}}
\toprule
Method         & OSOP             & ZS6D      & MegaPose               & GenFlow           \\ \midrule
$\#$ Ref. Views & \textgreater{}1K & 300       & 576                    & \textgreater{}100 \\ \midrule \midrule
Method         & GigaPose         & FoundPose & RayPose                & Ours              \\ \midrule
$\#$ Ref. Views  & 162              & 798       & 16 (8 coarse + 8 fine) & \cellcolor[HTML]{EFEFEF}\textbf{10}       \\ \bottomrule
\end{tabular}
}
\caption{Comparison of the number of reference views.}
\label{Tab: reference image number}
\end{table}

\paragraph{Dataset Comparisons and Ablations.}
Table~\ref{Tab: dataset information comparison} compares our dataset with the synthetic dataset from MegaPose~\cite{megapose}. Our dataset contains a larger number of object assets, more query images spanning diverse scenarios, and a wider range of object data sources.

In Table~\ref{Tab: dataset ablation}, we analyze the effect of training for 10k iterations under different dataset settings. The results show that our dataset consistently yields better performance than the MegaPose dataset. We further conduct ablations of our dataset, which verify the effectiveness of our data construction across different scenarios.

\begin{table}[t]
\renewcommand{\arraystretch}{1.3}
\resizebox{\columnwidth}{!}{\begin{tabular}{@{}c|ccc@{}}
\toprule
Dataset & \# Objects & \# Query Images & Object Sources \\ \midrule
MegaPose 2M & 53k & 2M & ShapeNet, Google-Scanned-Objects \\
Ours & \cellcolor[HTML]{EFEFEF}\textbf{190k} & \cellcolor[HTML]{EFEFEF}\textbf{30M} & Objaverse, Toys4K, 3D-FUTURE, ABO, HSSD \\ \bottomrule
\end{tabular}
}
\caption{Comparison of dataset scale and object sources.}
\label{Tab: dataset information comparison}
\end{table}

\begin{table}[t]
\renewcommand{\arraystretch}{1}
\resizebox{\columnwidth}{!}{\begin{tabular}{@{}ccccc|c@{}}
\toprule
          & \multicolumn{4}{c|}{Our dataset scenarios} &      \\ \cmidrule(lr){2-5}
\multirow{-2}{*}[3pt]{MegaPose 2M} & Centric   & Uncentric & EnvMap    & Edited    & \multirow{-2}{*}[6pt]{\begin{tabular}[c]{@{}c@{}}AR\\ (LM-O)\end{tabular}} \\ \midrule 
\ding{51} & \ding{55}    & \ding{55}   & \ding{55}   & \ding{55}   & 28.5 \\
\ding{55}                     & \ding{51} & \ding{51} & \ding{51} & \ding{51} & \cellcolor[HTML]{EFEFEF}\textbf{34.3}                                 \\
\ding{51} & \ding{51}    & \ding{51}   & \ding{51}   & \ding{51}   & 34.2 \\ \midrule \midrule
          & \ding{51}    & \ding{55}   & \ding{55}   & \ding{55}   & 22.6 \\
          & \ding{51}    & \ding{51}   & \ding{55}   & \ding{55}   & 30.8 \\
          & \ding{51}    & \ding{51}   & \ding{51}   & \ding{55}   & 33.6 \\
\multirow{-4}{*}[0pt]{---}           & \ding{51} & \ding{51} & \ding{51} & \ding{51} & \cellcolor[HTML]{EFEFEF}\textbf{34.3}                                 \\ \bottomrule
\end{tabular}
}
\caption{Performance under different training data settings. ``Centric", ``Uncentric", ``EnvMap", and ``Edited" correspond to the four components of our dataset described in Sec.~\ref{subsec: image data} of the main paper, respectively. Evaluation is conducted on the LM-O~\cite{brachmann2014learning} subset.}
\label{Tab: dataset ablation}
\end{table}

\paragraph{Comparisons with Refinement Networks.}
Table~\ref{Tab: compare with refine} presents performance comparisons after applying the MegaPose refinement network~\cite{megapose}. Specifically, we use the pose estimates reported in Table~\ref{tab:compare} of the main paper for each method as coarse initializations, which are then refined using the refinement network. As shown in the table, our method benefits much from the additional refinement. Owing to its accurate initial pose estimates, our method achieves performance that remains competitive with other approaches.

\begin{table}[t]
\renewcommand{\arraystretch}{1}
\resizebox{\columnwidth}{!}{\begin{tabular}{@{}lcccccc@{}}
\toprule
\multicolumn{1}{l|}{Method} & LM-O & T-LESS & TUD-L & IC-BIN & YCB-V & Average \\ \midrule
\multicolumn{1}{l|}{Ours (coarse)} & 43.0 & 34.1 & 56.8 & 24.3 & 47.4 & 41.1 \\ \midrule
\multicolumn{1}{l|}{MegaPose} & 49.9 & 47.7 & 65.3 & 36.7 & 60.1 & 51.9 \\
\multicolumn{1}{l|}{GigaPose} & 55.6 & \cellcolor[HTML]{EFEFEF}\textbf{54.6} & 57.8 & \cellcolor[HTML]{EFEFEF}\textbf{44.3} & \underline{63.4} & 55.1 \\
\multicolumn{1}{l|}{FoundPose} & 55.7 & 51.0 & 63.3 & 43.3 & \cellcolor[HTML]{EFEFEF}\textbf{66.1} & 55.9 \\
\multicolumn{1}{l|}{RayPose} & \underline{56.2} & \underline{53.8} & \cellcolor[HTML]{EFEFEF}\textbf{66.5} & 41.6 & 62.8 & \underline{56.2} \\
\multicolumn{1}{l|}{Ours} & \cellcolor[HTML]{EFEFEF}\textbf{57.4} & 52.1 & \underline{65.7} & \underline{44.1} & 63.3 & \cellcolor[HTML]{EFEFEF}\textbf{56.5} \\ \bottomrule
\end{tabular}
}
\caption{Performance after applying refinement network.}
\label{Tab: compare with refine}
\end{table}

\section{Additional Qualitative Results}
\label{app: visual}

In Fig.~\ref{fig:comparison2}, we present additional qualitative results, comparing our method with existing approaches.

\end{document}